\documentclass[letterpaper]{article} 
\usepackage{aaai2027}  
\usepackage[hyphens]{url}  
\usepackage{graphicx} 
\usepackage{natbib}  
\usepackage{caption} 
\usepackage{algorithm}
\usepackage{algorithmic}

\usepackage{newfloat}
\usepackage{listings}
\DeclareCaptionStyle{ruled}{labelfont=normalfont,labelsep=colon,strut=off} 
\floatstyle{ruled}
\newfloat{listing}{tb}{lst}{}
\floatname{listing}{Listing}

\usepackage{booktabs}
\usepackage{xcolor}
\usepackage{tcolorbox}
\tcbuselibrary{breakable}

\definecolor{workedAgentBg}{HTML}{EAF3FF}
\definecolor{workedResultBg}{HTML}{F2F4F7}
\definecolor{workedFailBg}{HTML}{FCE8E8}
\definecolor{workedPassBg}{HTML}{E7F2FF}
\definecolor{workedFailFrame}{HTML}{B94A58}
\definecolor{workedMemoryFrame}{HTML}{A87513}
\definecolor{workedEnvProbeFrame}{HTML}{2869A8}

\newtcolorbox{workedshared}[2]{%
    breakable,
    colback=#1,
    colframe=#1!65!black,
    title={#2},
    fonttitle=\small\sffamily\bfseries,
    fontupper=\small,
    boxrule=0.45pt,
    arc=1mm,
    left=4pt,
    right=4pt,
    top=3pt,
    bottom=3pt,
    before skip=4pt,
    after skip=4pt,
    before upper={\raggedright\setlength{\parindent}{0pt}\setlength{\parskip}{0pt}%
        \setlength{\emergencystretch}{3em}}%
}
\newtcolorbox{workedlaneheader}[2]{%
    colback=#1!8!white,
    colframe=#1,
    colbacktitle=#1,
    coltitle=white,
    title={#2},
    fonttitle=\scriptsize\sffamily\bfseries,
    fontupper=\scriptsize\ttfamily,
    boxrule=0.8pt,
    arc=1mm,
    left=3pt,
    right=3pt,
    top=2pt,
    bottom=2pt,
    before skip=1pt,
    after skip=2pt,
    before upper={\raggedright\setlength{\parindent}{0pt}\setlength{\parskip}{0pt}%
        \setlength{\emergencystretch}{3em}}%
}
\newtcolorbox{workedagentbubble}[2]{%
    breakable,
    width=0.89\linewidth,
    flush right,
    colback=workedAgentBg,
    colframe=#1,
    colbacktitle=#1,
    coltitle=white,
    title={#2},
    fonttitle=\scriptsize\sffamily\bfseries,
    fontupper=\scriptsize\ttfamily,
    boxrule=0.55pt,
    arc=0.8mm,
    left=2.5pt,
    right=2.5pt,
    top=2pt,
    bottom=2pt,
    before skip=1pt,
    after skip=1pt,
    before upper={\raggedright\setlength{\parindent}{0pt}\setlength{\parskip}{0pt}%
        \setlength{\emergencystretch}{3em}}%
}
\newtcolorbox{workedsystembubble}[3]{%
    breakable,
    width=0.89\linewidth,
    flush left,
    colback=#1,
    colframe=#2,
    colbacktitle=#2,
    coltitle=white,
    title={#3},
    fonttitle=\scriptsize\sffamily\bfseries,
    fontupper=\scriptsize\ttfamily,
    boxrule=0.55pt,
    arc=0.8mm,
    left=2.5pt,
    right=2.5pt,
    top=2pt,
    bottom=2pt,
    before skip=1pt,
    after skip=1pt,
    before upper={\raggedright\setlength{\parindent}{0pt}\setlength{\parskip}{0pt}%
        \setlength{\emergencystretch}{3em}}%
}
\newtcolorbox{workedlanesection}[1]{%
    colback=#1!8!white,
    colframe=#1,
    fontupper=\small\sffamily\bfseries,
    boxrule=0.6pt,
    arc=0.8mm,
    left=4pt,
    right=4pt,
    top=2pt,
    bottom=2pt,
    before skip=3pt,
    after skip=2pt
}

\title{Grounding Agent Memory: Environment-Probing Curation for Enterprise Agents}
\author{
    Susheel Suresh\thanks{Correspondence to \texttt{sussuresh@microsoft.com}.},
    Hazel Mak,
    Sahil Bhatnagar,\\
    Chhaya Methani,
    Alejandro Gutierrez Munoz
}
\affiliations{
    Microsoft Corporation\\
    One Microsoft Way, Redmond, WA 98052, USA
}

\usepackage[colorlinks=true,citecolor=blue,linkcolor=black,urlcolor=blue,breaklinks=true]{hyperref}

\begin{document}

\maketitle

\begin{abstract}
    Persistent memory is entering production-oriented agent platforms to help
long-horizon agents accumulate experience across sessions.  Yet a post-task
curator agent restricted to completed trajectories can preserve errors,
overgeneralize partial evidence, or retain stale knowledge.  We introduce
\emph{environment-probing curation}, a deployment-compatible extension that
gives an existing asynchronous curator agent least-privilege, read-only world
tools to check, scope, and refresh candidate memories.  It requires no model
retraining and leaves the task agent, retriever, memory representation, and
production write authority unchanged.  In a production-like GitHub Copilot
(GHCP) harness built on its SDK, we compare stateless execution, full
in-context learning, \emph{GHCP + Mem}, and \emph{GHCP + Mem (w/ Env
Probing)} on CLBench database exploration and 90 adapted APEX
management-consulting tasks.  On CLBench, probing raises pass rate from 39\%
to 73\% and pass-discounted reward from 8.60 to 22.60 while reducing queries
from 8.8 to 4.7 per question and task-agent cost from \$3.38 to \$1.68.
Across six APEX worlds, all 18 memory-versus-baseline mean reward comparisons
are positive and task-agent tool calls fall by 16--75\%; probing gives the
best task-agent reward gain per dollar in five worlds.  Probing also attains
higher mean reward than GHCP + Mem on both Sonnet 4.6 and Opus 4.7 without
schema drift.
Environment probing therefore turns existing agent-memory curation into an
environment-informed, auditable process while preserving a compact task-time
interface.
\end{abstract}

\section{Introduction}
\label{sec:intro}
    Large language model (LLM) agents are moving beyond bounded task execution
toward sustained work across sessions: implementing features in evolving
codebases, analyzing enterprise data, and supporting customers through
external applications.  Success depends on accumulating feedback and,
especially, knowledge of the latent environment structure shared across
related tasks.  Stateless execution discards trajectories, observations,
discoveries, and missteps at every session boundary, preventing experience
from improving subsequent work
\citep{asawa2026clbench,he2026memoryarenabenchmarkingagentmemory}.

A natural remedy is external memory, with three core operations: create
records from prior interactions, update them as evidence accumulates, and
recall relevant records during later execution
\citep{hu2026memoryageaiagents,park2023generative,packer2023memgpt,
zhang2025ace,ouyang2025reasoningbank}.  Designs range from full-context
trajectory replay, per-trajectory summaries, and a mutable notepad to
queryable indexes.  Structured records, relational encodings, hierarchical
tiers, and reasoning-aware retrieval make the latter increasingly capable
\citep{chhikara2025mem0,xu2025amem,kang2025memoryos,shu2026remem,
ji2026graphmemory}.  These ideas are entering practical agent stacks:
Claude Managed Agents\footnote{Claude Managed Agents Memory:
\url{https://platform.claude.com/docs/en/managed-agents/memory}.}
and Microsoft Copilot Studio\footnote{Copilot Studio
Memory (preview): \url{https://learn.microsoft.com/en-us/microsoft-copilot-studio/agents-experience/memory-overview}.}
expose memory features,
while Mem0 and Zep support production-oriented memory workflows
\citep{anthropic2026managedagentsmemory,
microsoft2026copilotstudiomemory,mem02026platform,zep2026platform}.

Persistent state alone, however, does not guarantee continual learning.
CLBench shows that memory can encode spurious generalizations and stale
beliefs under environment drift \citep{asawa2026clbench};
\citet{xiong2026memorymanagement} identify error propagation and misaligned
experience replay.  Post-task curation in Mem0, Claude Managed Agents
Dreams, ACE, ReasoningBank, and ReMe operates over some combination of existing
records, completed trajectories, feedback, and usage signals
\citep{chhikara2025mem0,anthropic2026managedagentsdreams,zhang2025ace,
ouyang2025reasoningbank,cao2026reme}.

This retrospective evidence boundary is fundamentally incomplete.  A
trajectory is a single, partial, and often mistake-laden observation of the
environment.  Its lemmas can (i)~memorize an instance answer rather than its
procedure, (ii)~inherit an inefficient path, (iii)~assert an unverifiable
scope, (iv)~leave blind spots in unvisited regions, or (v)~go stale as the
world changes.  Deferring verification to task time forces the responding
agent to spend scarce tool calls rechecking uncertain memories rather than
directly solving the current task.

We therefore propose \emph{environment-probing curation}.  After each task,
the harness instantiates a post-task curator agent with memory
CRUD; it receives the completed trajectory and grade after a
non-writing distillation step, then uses read-only world tools to check
candidate claims, test their scope across omitted states,
re-enact procedures, and refresh stale entries before writing.  This is
consistent with constructive accounts of episodic memory, where prior
experience is recombined to simulate possible and counterfactual events
\citep{bartlett1932remembering,schacter2007constructive,
schacter2012future,schacter2015futurecounterfactual}, and with interactive
learning, where acting in an environment yields competence unavailable from
passive observation alone \citep{sukhbaatar2018asymmetric}.

\begin{figure}[t]
    \centering
    \includegraphics[width=\columnwidth]{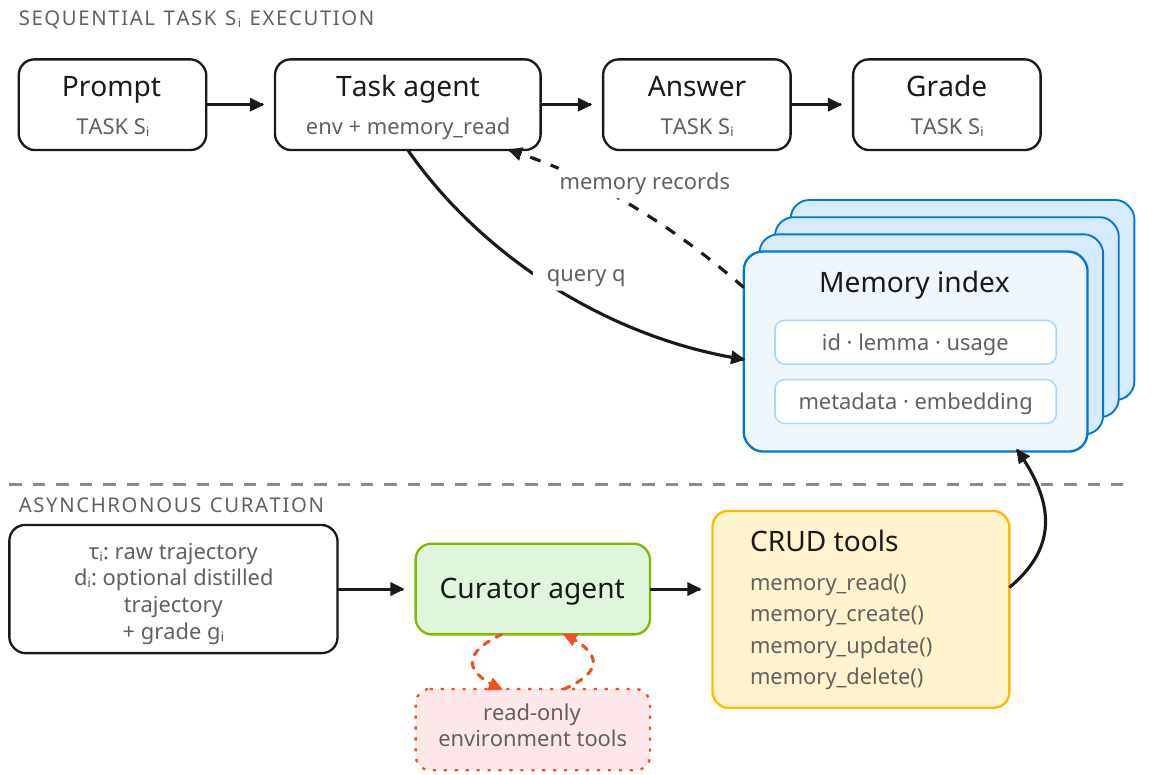}
    \caption{Agent roles and tool boundaries for task \(S_i\).  The horizontal
    dashed line separates sequential task execution from asynchronous
    curation.  The task agent uses environment tools and read-only memory.
    After task closure, the curator agent receives raw trajectory \(\tau_i\),
    distilled trajectory \(d_i\), and the grade; only the curator
    agent can write memory.  The red dashed loop marks read-only probing.}
    \label{fig:curator-env-probing}
\end{figure}

The mechanism is readily incorporated into existing enterprise agents.  It
requires no retraining and leaves the responding agent, retriever, record
representation, and production write authority unchanged.  The asynchronous
curator agent receives only a least-privilege, read-only subset of existing
connectors or MCP tools
\citep{microsoft2026copilotstudioconnectors,microsoft2026copilotstudiomcp}.
Probes remain off the user-facing critical path and task budget, inherit
platform authentication and auditing, and disappear without a safe read
surface.  Figure~\ref{fig:curator-env-probing} isolates this boundary: only the
asynchronous curator agent gains world tools.  With the task-time memory
interface and CRUD lifecycle fixed, incremental gains measure write-time
evidence quality rather than added task-agent capacity.

In our production-like GitHub Copilot (GHCP) harness, built on its SDK, each
task gets a fresh agent session while a persistent index spans sessions and
external tools expose the environment \citep{github2026copilotsdk}.  Both
benchmarks mirror deployed enterprise work: CLBench models data analysis over
evolving organization-specific databases; adapted APEX contributes 90
consulting-analyst tasks across six heterogeneous document worlds
\citep{asawa2026clbench,vidgen2026apex}.  Environment probing raises CLBench
pass rate from 39\% to 73\%, reward from
8.60 to 22.60, and cuts task-agent cost from \$3.38 to \$1.68.  Across APEX,
all 18 memory-versus-baseline mean reward comparisons are positive and
probing gives the best task-agent reward gain per dollar in five of six
worlds.  Our contributions are a diagnosis of trajectory-only
generalization failure, a deployment-compatible
\emph{propose--probe--commit} curator agent, and a cost-aware evaluation
showing that environment-informed procedures improve correctness while eliminating
repeated environment exploration.

\section{Related Work}
\label{sec:related}
    Agent memory extends retrieval-augmented generation from external knowledge
corpora to experience accumulated by the agent itself
\citep{lewis2020rag,hu2026memoryageaiagents}.  We focus on prompt-based
systems that leave model weights fixed and organize the literature by what
they aim to transfer: persistent facts about a user or environment, and
procedures learned from prior execution.

\paragraph{Factual memory.}
Early systems establish the basic create--store--recall lifecycle.
Generative Agents retrieves episodes by recency, importance, and relevance,
while MemGPT exposes OS-like memory tiers that the agent manages through
tools \citep{park2023generative,packer2023memgpt}.  Later work strengthens
creation and maintenance: MemoryBank summarizes dialogue with forgetting,
Mem0 reconciles new records through explicit CRUD decisions, MemoryOS
separates short-, mid-, and long-term stores, and SeCom chooses coherent
segments as the memory unit
\citep{zhong2024memorybank,chhikara2025mem0,kang2025memoryos,pan2025secom}.
SimpleMem jointly filters low-density dialogue, normalizes temporal and
referential content, and adapts retrieval across semantic, lexical, and
symbolic indexes \citep{liu2026simplemem}.  Graph systems replace isolated
records with linked episodic, semantic, temporal, or provenance-aware
structures, improving multi-hop recall and stale-fact invalidation
\citep{xu2025amem,gutierrez2024hipporag,anokhin2024arigraph,
rasmussen2025zep,ji2026graphmemory,shu2026remem}.
PlugMem bridges the factual and procedural classes with provenance-linked
records and routed retrieval \citep{yang2026plugmem}.

\paragraph{Procedural memory.}
Procedural systems distill behavior that can improve a later task.  Reflexion
writes verbal lessons from feedback, Synapse retrieves successful trajectory
exemplars, and ExpeL contrasts successes and failures to extract transferable
insights \citep{shinn2023reflexion,zheng2024synapse,zhao2024expel}.
Contextual replay buffers, dynamic cheatsheets, and reusable reasoning
templates compress prior execution at different granularities
\citep{liu2025contextual,suzgun2025dynamic,yang2024bufferthoughts}.
Voyager turns environment feedback into executable skills, while Agent
Workflow Memory induces retrievable workflows
\citep{wang2023voyager,wang2024workflow}.  Recent systems make curation more
deliberate: MemP applies CRUD updates from execution feedback, ACE evolves
playbooks through generation and reflection, ReasoningBank distills
strategies from both successful and failed attempts, and ReMe adds
validation, deduplication, utility pruning, and task-conditioned rewriting
\citep{fang2026memp,zhang2025ace,ouyang2025reasoningbank,cao2026reme}.

\paragraph{Evidence boundary.}
These advances change memory content, representation, retrieval, or
rewriting, but post-task curation still operates mainly over existing records,
recorded trajectories, grades, and usage signals.  Consequently, stronger
reflection cannot recover states the task policy never observed or determine
whether a trajectory-derived rule remains true after environment drift.  Our
contribution is orthogonal: read-only world tools let either a factual or
procedural curator agent independently check a claim, re-enact a procedure,
inspect omitted states, test scope, and refresh stale knowledge before
writing.

\section{Method}
\label{sec:method}
    \subsection{Online Setting and Task Agent}

Let \(\mathcal{S}=(S_1,\ldots,S_N)\) be an online stream of related tasks in
an environment whose state is \(E_i\) when task \(S_i\) arrives.  The
environment can evolve as \(E_{i+1}\sim\Delta(E_i)\), including changes that
are not announced to the agent.  Future tasks are hidden, task \(S_i\) must
close before \(S_{i+1}\) is revealed, and no task is revisited.  Model
parameters \(\theta\) remain fixed and every task starts a fresh session.

The \emph{task agent} \(\mathcal{A}_{\theta}\) is this fresh LLM session.  Its
sole goal is to solve \(S_i\).  It receives the task's environment tools
\(\mathcal{T}_{E_i}\), such as database queries in CLBench or document,
analysis, and artifact tools in APEX.  In the stateless setting, no shared
state is available to the agent while it solves successive tasks: each new
session begins without access to earlier trajectories or discoveries.

In the stateful setting, an external memory store
\(\mathcal{M}_{i-1}\) persists across sessions.  The task agent has read-only
access through \texttt{memory\_read}; the retriever
\(\mathcal{R}(\cdot,\mathcal{M}_{i-1})\) returns a small set of records
relevant to the task agent's request.
The agent has no memory create, update, or delete tool, so it cannot change
shared memory during task execution.  We write
\begin{equation}
  \tau_i=\mathcal{A}_{\theta}
  (S_i;\mathcal{T}_{E_i},\mathcal{R}(\cdot,\mathcal{M}_{i-1})),
  \label{eq:main-task-agent}
\end{equation}
with the retrieval argument omitted in the stateless setting.  The raw
trajectory \(\tau_i\) contains the request, memory reads, action--observation
pairs, and submitted answer.  Terminal feedback \(g_i\) arrives only after
the task closes.  The task-agent model, prompt, and environment tools are
fixed across the memory conditions.

\subsection{Post-Task Memory Curation}

We now turn from reading memory during a task to creating and updating memory
after it.  The \emph{curator agent} \(\mathcal{C}_{\phi}\) is a separate LLM
agent instantiated after the task agent submits its answer and \(g_i\)
becomes available.  It is not a continuation of the task-agent session, does
not answer \(S_i\), and cannot see future tasks.  It receives the completed
trajectory, terminal feedback, and relevant existing records.

The curator agent has four memory tools: \texttt{memory\_read},
\texttt{memory\_create}, \texttt{memory\_update}, and
\texttt{memory\_delete}.  It is the only agent allowed to mutate
\(\mathcal{M}\).  At this stage of the method, the curator agent has no live
task-environment tools; it reasons only over the completed-task evidence and
the memory store.  It can retrieve related records before writing, so new
evidence is reconciled with existing memory rather than appended blindly.
Each record carries a category, confidence, applicability scope, concise
lemma, provenance, utility, and usage metadata.
After curation, the committed store becomes \(\mathcal{M}_i\) and is
available when \(S_{i+1}\) begins.

\paragraph{Trajectory distillation.}
We apply the non-writing preprocessing transformation
\begin{equation}
  d_i=\mathcal{D}_{\psi}(\tau_i).
  \label{eq:main-distillation}
\end{equation}
The distiller transforms the full raw trajectory \(\tau_i\) into the
distilled trajectory \(d_i\).  It retains the task, retrieved memories,
decisive observations, procedures, unresolved assumptions, and answer, but
receives no terminal feedback, memory tools, or environment tools and cannot
write memory.  Appendix~\ref{app:distiller-prompt} gives its prompt.  The
distilled trajectory \(d_i\) provides the compact primary view while
\(\tau_i\) remains available as raw evidence.

The system prompt for the curator agent in
Appendix~\ref{app:curator-prompt} gives it one goal: maintain a small set of
reliable, transferable, and actionable records that help future task agents
solve related tasks with fewer task-agent tool calls.  It instructs the curator
agent to store reusable facts, procedures, relations, tool conventions, and
scoped warnings rather than task answers or incidental values.  Before
writing, it considers evidential support, transfer value, scope,
actionability, and overlap, then can create a record, update or merge one,
narrow its scope, or delete it.  It skips unsupported, redundant, trivial, or
task-specific content, and a passing grade does not automatically validate
every intermediate assumption.

The shared user-message template
(Appendix~\ref{app:curator-user-message}) supplies the raw trajectory
\(\tau_i\), distilled trajectory \(d_i\), terminal feedback \(g_i\),
and related records.  The curator agent finishes before \(S_{i+1}\) is
revealed, so future tasks cannot leak into memory and partial writes cannot
enter task execution.

\subsection{Environment-Probing Curation}

The left column of Figure~\ref{fig:clbench-curator-memory-records} in
Appendix~\ref{app:clbench-memory-record-examples} presents representative
records produced by the trajectory-only curator agent during the CLBench
database-exploration runs.  They make the retrospective evidence boundary
described in Section~\ref{sec:intro} concrete: each source trajectory is a
single, partial, and potentially mistake-laden observation.  Consequently,
one record preserves an incorrect aggregation and its answer without
supplying the replacement procedure, another gives only a broad domain map
while omitting the needed relation, and a third retains the removed
\texttt{attrs\_g3} name after schema drift.  In each case, memory transfers
some prior knowledge but leaves the next task agent to establish whether it
is actionable and current.

Environment-probing curation addresses this evidence boundary.  It keeps the
task agent, retriever, distillation setting, curator agent, memory schema, and
CRUD policy unchanged.  Only after task closure does the pipeline give the
curator agent a safe, read-only subset of the environment tools and a short
instruction to probe when a candidate or existing record is uncertain.
Figure~\ref{fig:curator-env-probing} shows this added tool loop, and
Appendix~\ref{app:probing-curator-prompt} shows the two additions to the
otherwise identical prompt for the curator agent.

The curator agent follows a \emph{propose--probe--commit} process.  After
proposing a candidate memory, it makes targeted read-only tool calls to
investigate specific uncertainties, then uses the observations to create,
revise, narrow, delete, or skip the record.  Concretely, it can
(i)~distinguish an incidental answer from a reusable relation, (ii)~compare
the observed procedure with a shorter path, (iii)~test a claimed relation on
another slice, (iv)~check a procedure's required preconditions, (v)~inspect
relevant states omitted by the task trajectory, or (vi)~re-query the current
environment when drift is suspected.
Our hypothesis is that this limited read-only interaction is an effective way
to produce environment-informed memory records.

In CLBench, probes inspect tables, join keys, encodings, or post-migration
fields.  In APEX, they inspect file locations, document relevance, workbook
contents, or tool conventions.  In both cases, a probe evaluates a proposed
memory; it does not solve a future task.  Probes cannot mutate the
environment, enter the task trajectory, consume the task agent's budget, or
expose future tasks or labels.  This read-only design avoids side effects and
requires less authority than giving an asynchronous curator agent production
write access.  If no safe read surface exists, the curator agent falls back
to trajectory-only curation.  The side-by-side examples in
Figure~\ref{fig:clbench-curator-memory-records}
(Appendix~\ref{app:clbench-memory-record-examples}) qualitatively illustrate
more actionable, environment-informed memory records.  We next describe
the experiments in Section~\ref{sec:setup} and report their results in
Section~\ref{sec:results}.

\section{Experiment Setup}
\label{sec:setup}
    \paragraph{Systems and benchmarks.}
We compare four GitHub Copilot (GHCP) systems: GHCP (No Memory), GHCP + Full
ICL, GHCP + Mem, and GHCP + Mem (w/ Env Probing).  Full ICL prepends prior
trajectories, while both
memory systems expose the same \texttt{memory\_read} interface; environment
probing adds only read-only tools for the curator agent and the corresponding
instructions.  CLBench uses two schedules \citep{asawa2026clbench}.  The
primary 40-question \emph{drift} schedule hides a SQLite schema that changes
after question 20, testing reuse and stale-memory repair.  The 30-question
\emph{no-drift} schedule fixes the schema, separating validation of stable
joins and encodings from migration recovery; it evaluates both memory systems
on Sonnet 4.6 and Opus 4.7 with paired no-memory baselines.  Both schedules
hide joins and mix timestamp and price encodings; drift also renames fields
and adds soft deletes.
Adapted APEX contributes 90 management-consulting tasks from six
shared document worlds, requiring PDF, XLSX, DOCX, and PPTX discovery,
quantitative analysis, and MCP-style tools \citep{vidgen2026apex}.  The
original benchmark and leaderboard are available at
\url{https://www.mercor.com/apex/apex-agents-leaderboard/}.

\paragraph{Protocol and metrics.}
The primary CLBench and APEX experiments use \texttt{gpt-5.4}; within each
experiment, the task agent, distiller, and curator agent share the same base
model.  CLBench uses five paired seeded runs for every configuration; APEX
uses five runs per stateful configuration and three stateless runs.  We
report run-level means with 95\% Student-\(t\) confidence intervals.  The
no-drift study holds the canonical 30-question order fixed and uses five
paired runs per model--memory comparison;
Appendix~\ref{app:experimental-details} details its uncertainty estimates.

For task \(i\), let \(p_i\in\{0,1\}\) be its binary pass score and let \(q_i\)
be the number of task-agent tool calls counted by the benchmark.  Its
pass-discounted reward is
\begin{equation}
    r_i=p_i\left(1-\frac{q_i}{B}\right).
    \label{eq:main-pass-discounted-reward}
\end{equation}
We set \(B=15\) SQL-query calls for CLBench and \(B=100\) Archipelago tool
calls for APEX.  For tasks with multiple rubric criteria, we use a strict pass:
\(p_i=1\) only when every criterion passes, and \(p_i=0\) otherwise.  Thus, a
failed task receives zero, while a passing task receives more reward when it
uses fewer task-agent tool calls.  We also report pass rate, tool calls,
tokens, and USD cost.  Memory-management calls are excluded from \(q_i\),
and task-agent cost excludes the separately tracked curation phase.
Appendix~\ref{app:experimental-details} gives the complete evaluation and
accounting details.

\section{Results}
\label{sec:results}
    \subsection{CLBench}

Table~\ref{tab:clbench-main-results}(a) shows that every memory configuration
improves both correctness and pass-discounted reward over GHCP (No Memory).
Pass rate rises from 39\% to 61--73\%, while total reward rises from 8.60 to
20.00--22.60, or 2.3--2.6\(\times\) the baseline.  This is not a
brute-force accuracy gain: queries fall from 8.8 to 3.0--5.6 per question and
task-agent cost falls from \$3.38 to \$1.68--\$2.01.  Memory therefore makes
the agent both more likely to pass and less likely to spend its budget
rediscovering the schema, encodings, and tool conventions already encountered
earlier in the stream.

\begin{table*}[t]
    \centering
    \small
    \renewcommand{\arraystretch}{1.08}
    \textbf{(a) 40-question GPT-5.4 schedule with schema drift}\par
    \smallskip
    \begin{tabular}{@{}lrrrrrr@{}}
        \toprule
        Configuration & Pass (\%) & Total reward &
        \shortstack{Queries per\\question} &
        \shortstack{Input\\tokens} &
        \shortstack{Output\\tokens} &
        \shortstack{Task-agent\\cost} \\
        \midrule
        \shortstack[l]{GHCP\\(No Memory)}
            & $39 \pm 4$ & $8.60 \pm 0.83$ & $8.8 \pm 0.3$
            & $3.14\mathrm{M} \pm 14.5\mathrm{K}$
            & $93.4\mathrm{K} \pm 9.3\mathrm{K}$
            & $\$3.38 \pm 0.70$ \\
        \addlinespace[2pt]
        \shortstack[l]{GHCP +\\Full ICL}
            & $61 \pm 11$ & $21.39 \pm 3.83$ & $3.0 \pm 0.2$
            & $5.42\mathrm{M} \pm 569.5\mathrm{K}$
            & $20.5\mathrm{K} \pm 4.6\mathrm{K}$
            & $\$2.01 \pm 0.23$ \\
        \addlinespace[2pt]
        \shortstack[l]{GHCP +\\Mem}
            & $70 \pm 16$ & $20.00 \pm 6.52$ & $5.6 \pm 1.4$
            & $2.13\mathrm{M} \pm 685.2\mathrm{K}$
            & $55.9\mathrm{K} \pm 22.1\mathrm{K}$
            & $\$1.99 \pm 0.65$ \\
        \addlinespace[2pt]
        \shortstack[l]{GHCP + Mem\\(w/ Env Probing)}
            & $\mathbf{73 \pm 5}$ & $\mathbf{22.60 \pm 2.07}$ & $4.7 \pm 0.3$
            & $\mathbf{1.69\mathrm{M} \pm 97.8\mathrm{K}}$
            & $45.4\mathrm{K} \pm 6.0\mathrm{K}$
            & $\mathbf{\$1.68 \pm 0.14}$ \\
        \bottomrule
    \end{tabular}

    \medskip
    \textbf{(b) 30-question cross-model schedule without schema drift}\par
    \smallskip
    \begin{tabular}{@{}llccc@{}}
        \toprule
        Model & Memory system &
        \shortstack{Paired no-memory\\mean reward} &
        \shortstack{With-memory\\mean reward} &
        Lift \\
        \midrule
        Opus 4.7 & GHCP + Mem
            & $0.444 \pm 0.050$ & $0.696 \pm 0.036$ & $+0.252$ \\
        Opus 4.7 & GHCP + Mem (w/ Env Probing)
            & $0.458 \pm 0.048$ & $\mathbf{0.721 \pm 0.062}$ & $\mathbf{+0.263}$ \\
        Sonnet 4.6 & GHCP + Mem
            & $0.322 \pm 0.054$ & $0.673 \pm 0.097$ & $+0.351$ \\
        Sonnet 4.6 & GHCP + Mem (w/ Env Probing)
            & $0.327 \pm 0.055$ & $\mathbf{0.748 \pm 0.030}$ & $\mathbf{+0.421}$ \\
        \bottomrule
    \end{tabular}
    \caption{CLBench results: (a) GPT-5.4 on 40-task drift (migration after
    task 20), reporting strict pass, total
    Equation~\ref{eq:main-pass-discounted-reward} reward, SQL queries/task,
    tokens, and task-agent cost as means \(\pm\) 95\% Student-\(t\)
    confidence intervals over five paired independent runs; (b) fixed-order
    30-task no drift, reporting paired baseline/memory mean reward
    \(\pm\) standard deviation over five independent runs and their
    difference; calls exclude memory and
    tokens/cost exclude distillation/curation; bold marks highest pass/reward
    and lowest input/cost in (a), and highest reward/lift per model in (b).}
    \label{tab:clbench-main-results}
\end{table*}

\paragraph{Retention without prompt growth.}
GHCP + Full ICL confirms that prior trajectories contain useful signal and
uses the fewest SQL queries, but consumes 5.42M input tokens because its
context grows with the stream.  GHCP + Mem retrieves compact lemmas
on demand, raises pass rate further to 70\%, and uses 2.13M input tokens.
Environment-probed memory reaches the highest pass rate and reward with only
1.69M input tokens.  The two memory designs thus retain reusable experience
without making task-time context proportional to deployment age.

\begin{figure*}[t]
    \centering
    \includegraphics[width=0.98\textwidth]{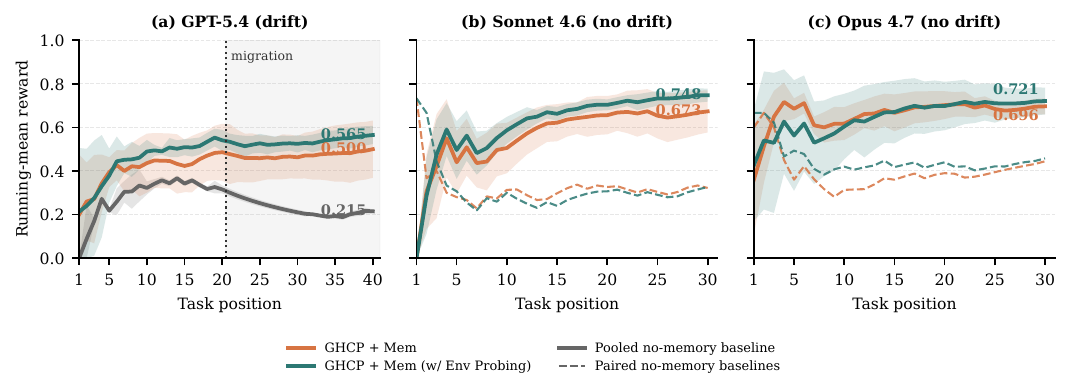}
    \caption{CLBench learning curves in the main evaluation.  Panel (a) shows
    GPT-5.4 on the 40-question drift schedule; the vertical marker denotes the
    migration after question 20, and the gray curve is the five-run paired
    baseline.  Panels (b) and (c) show Sonnet 4.6 and Opus 4.7 on the
    30-question no-drift schedule; same-color dashed lines are the paired
    no-memory rollouts.  Solid curves are means over five stateful runs and
    bands show one standard deviation.  Endpoints reproduce
    Table~\ref{tab:clbench-main-results}.}
    \label{fig:clbench-learning-main}
\end{figure*}

Figure~\ref{fig:clbench-learning-main}(a) shows when the gains accrue.
Environment probing already leads trajectory-only memory at the migration
boundary (0.541 versus 0.486 cumulative reward) and finishes at 0.565 versus
0.500; no memory ends at 0.215.  The persistent post-migration lead is
consistent with curator-side probes refreshing schema lemmas before later
task sessions retrieve them, rather than making every task agent detect and
repair drift independently.

\paragraph{No-drift cross-model results.}
Table~\ref{tab:clbench-main-results}(b) isolates memory from schema repair and
shows gains across Sonnet 4.6 and Opus 4.7.  GHCP + Mem improves over its
paired baseline by 0.351 and 0.252, while environment-probed memory improves
by 0.421 and 0.263, respectively.  Probing achieves the highest mean reward
on both models (0.748 and 0.721), and both memory systems finish above their
paired no-memory curves in Figure~\ref{fig:clbench-learning-main}(b--c).

\subsection{Adapted APEX}

Table~\ref{tab:apex-results} compares reward gain per task-agent dollar, with
the underlying reward gain and cost shown in each cell.  All 18 gains---six
worlds by three memory systems---are positive.  Appendix
\ref{app:extended-results} reports absolute reward and task-agent tool calls.

\begin{table}[t]
    \centering
    \footnotesize
    \setlength{\tabcolsep}{2.3pt}
    \renewcommand{\arraystretch}{1.03}
    \begin{tabular}{@{}lccc@{}}
        \toprule
        World [tier, \(N\)] &
        \shortstack{Full\\ICL} &
        Mem &
        Probe \\
        \midrule
        \shortstack[l]{\texttt{941eba66}\\{[}Easy, 15{]}}
            & \shortstack{\(0.410/\$\)\\\(+7.40,\ \$18.04\)}
            & \shortstack{\(0.858/\$\)\\\(+7.44,\ \$8.67\)}
            & \shortstack{\(\mathbf{0.991/\$}\)\\\(+7.40,\ \$7.47\)} \\
        \midrule
        \shortstack[l]{\texttt{d6c01a12}\\{[}Easy, 11{]}}
            & \shortstack{\(0.047/\$\)\\\(+0.58,\ \$12.52\)}
            & \shortstack{\(0.125/\$\)\\\(+1.00,\ \$7.95\)}
            & \shortstack{\(\mathbf{0.142/\$}\)\\\(+1.23,\ \$8.66\)} \\
        \midrule
        \shortstack[l]{\texttt{2a87e5cb}\\{[}Medium, 18{]}}
            & \shortstack{\(0.108/\$\)\\\(+2.64,\ \$24.42\)}
            & \shortstack{\(0.038/\$\)\\\(+0.58,\ \$15.36\)}
            & \shortstack{\(\mathbf{0.184/\$}\)\\\(+2.35,\ \$12.79\)} \\
        \midrule
        \shortstack[l]{\texttt{2f84c98b}\\{[}Medium, 17{]}}
            & \shortstack{\(0.026/\$\)\\\(+0.88,\ \$33.79\)}
            & \shortstack{\(0.110/\$\)\\\(+1.99,\ \$18.17\)}
            & \shortstack{\(\mathbf{0.189/\$}\)\\\(+3.08,\ \$16.28\)} \\
        \midrule
        \shortstack[l]{\texttt{d1b705c7}\\{[}Hard, 15{]}}
            & \shortstack{\(0.116/\$\)\\\(+3.45,\ \$29.78\)}
            & \shortstack{\(0.140/\$\)\\\(+2.36,\ \$16.93\)}
            & \shortstack{\(\mathbf{0.172/\$}\)\\\(+2.46,\ \$14.24\)} \\
        \midrule
        \shortstack[l]{\texttt{075ef4df}\\{[}Hard, 14{]}}
            & \shortstack{\(0.103/\$\)\\\(+1.47,\ \$14.34\)}
            & \shortstack{\(\mathbf{0.274/\$}\)\\\(+1.88,\ \$6.88\)}
            & \shortstack{\(0.271/\$\)\\\(+2.13,\ \$7.89\)} \\
        \bottomrule
    \end{tabular}
    \caption{APEX cost-adjusted efficiency (90 tasks): each world
    [difficulty, \(N\)] cell shows pass-discounted reward gain over a
    three-run no-memory baseline per dollar of candidate task-agent cost
    (top), then gain and mean cost/run over five candidate runs (bottom); costs exclude
    distillation/curation, calculations use unrounded means, and bold marks
    the highest ratio per world.}
    \label{tab:apex-results}
\end{table}

The largest reduction occurs where baseline discovery is most expensive:
world \texttt{941eba66} falls from 71.6 tool calls to 17.7--19.3, while the
indexed-memory systems reduce input consumption from 53.92M tokens to
6.56--7.67M and cost from \$54.30 to \$7--\$9 per run.  GHCP + Mem (w/ Env
Probing) achieves the best
task-agent reward gain per dollar in five of six worlds
(Table~\ref{tab:apex-results}); GHCP + Mem is marginally better in the
remaining world.  Full ICL can win raw reward in individual worlds, but its
growing context makes those gains expensive and even costs more than the
stateless baseline in one world.

Across APEX difficulty levels, memory can preserve an answer with fewer tool calls,
expose additional rubric evidence, or turn failure into success by leaving
budget for the final computation.  Tool reduction can therefore enable
correctness, not just lower latency.

\subsection{Why Memory and Probing Work}

\paragraph{Memory amortizes environmental discovery.}
On drift CLBench, GHCP + Mem raises pass rate from 39\% to 70\% and total
reward from 8.60 to 20.00 while reducing queries from 8.8 to 5.6 per task
(Table~\ref{tab:clbench-main-results}).  Across APEX, all 18
system-versus-baseline reward gains are positive, while task-agent calls fall
from baseline means of 30.0--71.6 to 13.9--28.4
(Table~\ref{tab:apex-results-detailed}).  Full ICL confirms that prior
trajectories contain reusable information, but requires 5.42M CLBench input
tokens, versus 2.13M for indexed memory and 1.69M with probing.  Indexed
records therefore preserve reusable schemas, relations, file maps, and
procedures without carrying the full interaction history.

\paragraph{Probing makes records more actionable.}
The two indexed-memory conditions retain the same task-time model, tools, and
read-only memory interface; probing provides additional environment evidence
during curation.  Relative to trajectory-only memory, probing raises drift
CLBench reward from 20.00 to 22.60, lowers queries from 5.6 to 4.7, and
reduces task-agent cost from \$1.99 to \$1.68.  Without drift, mean reward
rises from 0.673 to 0.748 on Sonnet and from 0.696 to 0.721 on Opus.

\paragraph{Qualitative analysis of memory records.}
Figure~\ref{fig:clbench-curator-memory-records} compares representative
records rather than one-to-one rewrites.  Trajectory-only curation records an
answer-anchored warning: ``Do not answer with
\texttt{AVG(items\_g2.prc\_usd)} over non-null rows; that produces about
52.96, but the benchmark's correct result is 96.23, so a different price
field and/or row subset is required.''  The probing curator instead records
an executable procedure: ``Join \texttt{items\_g2} to \texttt{taxn\_g2} on
\texttt{ref\_id}; filter \texttt{cat\_lvl=1} and the exact
\texttt{cat\_nm}; keep \texttt{items\_g2.prc>0}; then compare against the
filtered \texttt{AVG(prc)}.''  The same shift appears in the other records:
a broad \texttt{g1}/\texttt{g2}/\texttt{g3} map is contrasted with an
explicit \texttt{ref\_id} join and aggregation grain, while stale ``Use
\texttt{attrs\_g3}'' becomes ``Use
\texttt{product\_attributes\_g3}'' with brand filters and grouping.  The
right column thus specifies what a later agent should execute---source table,
join key, filters, grain, and current schema---rather than only what failed
or where to search.  Appendix~\ref{app:clbench-memory-record-examples} gives
the complete records; Appendices~\ref{app:worked-clbench} and
\ref{app:worked-apex} connect them to task-time behavior.

\paragraph{The effect depends on the remaining evidence gap.}
Probing improves over trajectory-only memory in five of six APEX worlds; the
largest increments are \(+1.77\) in \texttt{2a87e5cb} and \(+1.09\) in
\texttt{2f84c98b}, while \texttt{941eba66} changes by \(-0.04\)
(Table~\ref{tab:apex-results}).  Its additional no-drift gain is also larger
on Sonnet (\(+0.075\)) than on Opus (\(+0.025\)).  This variation is
consistent with probing being most useful when a trajectory leaves a join,
workbook location, or procedure unresolved, while adding little when the
trajectory already supports an actionable record.  Because uncertainty
intervals overlap, we treat this as a mechanism interpretation rather than a
resolved subgroup effect.

\paragraph{Matched trajectories connect records to behavior.}
In the matched CLBench task, no memory uses seven queries and fails,
trajectory-only memory uses nine and passes, and probing supplies a validated
\texttt{ref\_id} relation and passes in two
(Appendix~\ref{app:worked-clbench}).  In hard APEX, no memory uses 96 calls
and fails both criteria; trajectory-only memory transfers the
revenue-per-head procedure and passes in 11, while a validated workbook map
reduces probing to six (Appendix~\ref{app:worked-apex}).  These selected cases
do not establish the aggregate effect, but they illustrate how reusable
computations and environment-informed maps replace task-time rediscovery with
direct execution.

\FloatBarrier

\section{Conclusion}
\label{sec:conclusion}
    Environment probing addresses a fundamental limit of post-task memory:
a trajectory alone cannot establish that a lesson is correct, general, or
current.  Read-only world tools let the curator check lessons before they
enter long-lived memory without expanding task-time capabilities.  This
isolates the improvement to write-time evidence quality rather than added
task-agent capacity.  Across CLBench and adapted APEX, probing improves reward
while reducing repeated environment interaction and task-agent cost; its
advantage persists across the GPT-5.4, Sonnet 4.6, and Opus 4.7 model
families.  For deployment, production stacks retain the model, task agent,
retriever, record schema, and asynchronous CRUD lifecycle; only the curator gains
least-privilege, read-only connector or MCP access.  Probes add no write
authority, remain off the critical path, and inherit platform authentication
and auditing.

\bibliography{aaai2027}

\onecolumn
\appendix
\raggedbottom

\section{Implementation Details and Extensions}
\label{app:formal-method}
\subsection{Runtime Architecture and Session Lifecycle}

\paragraph{GitHub Copilot harness.}
We implement the runtime with the Python GitHub Copilot SDK, which exposes the
same agent engine as GitHub Copilot CLI.  Each containerized sandbox launches
a headless Copilot CLI process in server mode, and the SDK communicates with
it through JSON-RPC \citep{github2026copilotsdk}.  An SDK session specifies
the model, system instructions, custom tools, MCP servers, permission policy,
and event callbacks.  The callbacks provide the ordered model messages, tool
calls, and observations used to construct each trajectory.

\paragraph{Per-task lifecycle.}
For task \(S_i\), the harness creates a fresh task-agent session with the
environment tools and \texttt{memory\_read}.  The external index persists
across tasks, but the task agent has no memory-write tool.  In the reported
runs, after the task closes, a fresh distiller session receives the raw
trajectory \(\tau_i\)---not terminal feedback---and produces the distilled
trajectory \(d_i\).  Once feedback \(g_i\) is available, another fresh
session instantiates the curator agent with \(d_i\), \(g_i\), and retrieved
nearby memory records; the staged raw trajectory remains available as
supporting evidence.  The curator agent receives \texttt{memory\_read},
\texttt{memory\_create},
\texttt{memory\_update}, and \texttt{memory\_delete}; curation completes
before \(S_{i+1}\) is exposed.

The trajectory-only curator agent receives no task-environment tools.  Its ordinary
sandbox file readers can inspect the staged trajectory but cannot query the
live benchmark world.  The environment-probing curator agent uses the same model,
inputs, memory CRUD tools, and core system prompt, while additionally
receiving two probe-specific instructions and the task's read-only tool
subset.  For CLBench this subset is the database query interface; for adapted
APEX it is supplied through the read-only MCP configuration.
Appendix~\ref{app:prompt-templates} gives the compact prompt specifications
and highlights the probing additions.

\subsection{Enterprise Deployment Mapping}

The SDK--server decomposition mirrors the production-oriented Microsoft
Copilot Studio runtime, where an agent is configured from instructions, a
selected model, knowledge and memory, callable tools and skills, and connected
agents \citep{microsoft2026copilotstudioagentsoverview}.  Connectors can wrap
enterprise APIs, while MCP servers expose callable tools and file-like
resources such as API responses and document contents
\citep{microsoft2026copilotstudioconnectors,
microsoft2026copilotstudiomcp}.  Structured databases, document corpora, and
enterprise applications can therefore all instantiate the environment-tool
interface used by our curator agent.

In such a deployment, probing does not require a second integration path.
The asynchronous curator agent can be assigned a least-privilege, read-only subset
of connectors or MCP tools already registered for the responding agent.
Existing authentication, authorization, and audit boundaries remain in
force, and the curator agent receives no production write authority.  This makes
the intervention compatible with increasingly common managed-memory
abstractions \citep{microsoft2026copilotstudiomemory} without retraining the
model or changing the task-time agent.

\section{Experimental and Evaluation Details}
\label{app:experimental-details}
\subsection{Benchmark Construction}

\paragraph{CLBench database exploration.}
The primary CLBench stream contains 40 SQL questions over a hidden SQLite
database \citep{asawa2026clbench}.  The agent must discover tables, joins,
encodings, and conventions through queries.  Format traps include prices in
dollars versus cents, epoch-millisecond versus ISO timestamps, and abbreviated
column names.  After question 20, an unannounced migration renames tables,
splits columns, and introduces soft deletes, testing both schema transfer and
repair of previously valid memory.  The cross-model study instead uses the
canonical 30-question, single-stage schedule without migration.

\paragraph{Adapted APEX Agents.}
APEX Agents contains 480 independent workplace tasks spanning management
consulting, law, and financial analysis \citep{vidgen2026apex}.  We select
the management-consulting subset and group questions by
\texttt{(domain, world\_id)}, turning each shared world into an ordered
continual-learning stream.  This yields six worlds and 90 questions, with
11--18 questions per world.  Tasks require discovery across PDF, XLSX, DOCX,
and PPTX files, including embedded images; quantitative analysis through code
execution; and artifact production through MCP-style Archipelago tools.
Grouping by world makes file locations, workbook layouts, tool conventions,
and distinctions among hard-negative documents reusable across tasks.

\subsection{Models and Run Protocol}

The primary 40-question CLBench and adapted APEX studies use
\texttt{gpt-5.4} at xhigh reasoning effort for sessions of the task agent,
distiller, and curator agent.  The no-drift study uses Sonnet 4.6 at high
effort and Opus 4.7 at xhigh effort for all applicable roles.  Memory conditions use
\texttt{intfloat/e5-base-v2} embeddings.

The primary CLBench study uses five independently shuffled, paired runs for
every configuration.  Adapted APEX uses five runs per stateful configuration
and three stateless runs.  Shuffles are seeded and shared across systems.
Reported intervals are 95\% Student-\(t\) intervals over run-level
aggregates.  The no-drift study instead holds the canonical 30-question order
fixed; each model--memory comparison has five paired memory and GHCP (No
Memory) runs, and uncertainty is the across-run standard deviation.

\subsection{Metrics and Accounting}

For task \(i\), let \(p_i\in\{0,1\}\) be its binary pass score and let \(q_i\)
be the number of task-agent tool calls counted by the benchmark.  CLBench has
one correctness criterion.  For tasks with multiple grader criteria, we use
a strict pass: \(p_i=1\) only when every criterion passes, and \(p_i=0\)
otherwise.  We compute the primary pass-discounted reward \(r_i\) using
Equation~\ref{eq:main-pass-discounted-reward}, with \(B=15\) exploratory SQL
queries for CLBench and \(B=100\) Archipelago tool calls for APEX.  For
trajectory diagnosis only, APEX also reports the fractional-criteria score
\(r_i^{\mathrm{frac}}=(k_i/K_i)(1-q_i/B)\), where \(k_i\) of \(K_i\)
criteria pass.  We report total reward as the sum of per-task rewards and
mean reward as that total divided by the number of tasks.

Memory-management and harness-internal calls are excluded from \(q_i\).
Reported tokens and USD cost cover the task-agent response phase; usage by the
distiller and curator agent is tracked separately.  In addition to reward, we
report pass rate, task-agent tool calls per question, and input/output tokens.
At task position \(j\), a learning-curve point is the across-run mean of the
running mean reward through position \(j\); shaded bands show one across-run
standard deviation.

\section{Per-World APEX Results}
\label{app:extended-results}
Table~\ref{tab:apex-results-detailed} reports the absolute rewards,
task-agent tool calls, and confidence intervals underlying the compact gains in
Table~\ref{tab:apex-results}.  This breakdown preserves the variation across
difficulty tiers and document environments summarized in
Section~\ref{sec:results}.

\begin{table}[htbp]
    \centering
    \small
    \renewcommand{\arraystretch}{1.08}
    \begin{tabular}{@{}llcccc@{}}
        \toprule
        World [tier, \(N\)] & Metric &
        \shortstack{GHCP\\(No Memory)} &
        \shortstack{GHCP +\\Full ICL} &
        \shortstack{GHCP +\\Mem} &
        \shortstack{GHCP + Mem\\(w/ Env Probing)} \\
        \midrule
        \shortstack[l]{\texttt{941eba66}\\{[}Easy, 15{]}}
            & Reward & \(1.12 \pm 0.52\) & \(8.51 \pm 1.58\)
            & \(\mathbf{8.56 \pm 0.64}\) & \(8.52 \pm 2.01\) \\
        & Tool calls & \(71.6 \pm 10.7\) & \(\mathbf{17.7 \pm 2.2}\)
            & \(19.3 \pm 1.1\) & \(19.3 \pm 2.4\) \\
        \midrule
        \shortstack[l]{\texttt{d6c01a12}\\{[}Easy, 11{]}}
            & Reward & \(2.30 \pm 0.67\) & \(2.88 \pm 1.01\)
            & \(3.29 \pm 0.51\) & \(\mathbf{3.52 \pm 0.50}\) \\
        & Tool calls & \(40.1 \pm 12.6\) & \(\mathbf{24.2 \pm 2.9}\)
            & \(25.8 \pm 1.7\) & \(25.0 \pm 2.1\) \\
        \midrule
        \shortstack[l]{\texttt{2a87e5cb}\\{[}Medium, 18{]}}
            & Reward & \(0.48 \pm 1.07\) & \(\mathbf{3.12 \pm 2.10}\)
            & \(1.06 \pm 0.88\) & \(2.83 \pm 1.53\) \\
        & Tool calls & \(33.5 \pm 8.9\) & \(\mathbf{13.9 \pm 2.3}\)
            & \(21.3 \pm 1.7\) & \(21.0 \pm 1.5\) \\
        \midrule
        \shortstack[l]{\texttt{2f84c98b}\\{[}Medium, 17{]}}
            & Reward & \(6.52 \pm 0.60\) & \(7.40 \pm 2.48\)
            & \(8.51 \pm 3.08\) & \(\mathbf{9.60 \pm 1.33}\) \\
        & Tool calls & \(30.0 \pm 8.4\) & \(\mathbf{18.5 \pm 4.7}\)
            & \(25.3 \pm 4.4\) & \(23.6 \pm 0.8\) \\
        \midrule
        \shortstack[l]{\texttt{d1b705c7}\\{[}Hard, 15{]}}
            & Reward & \(2.02 \pm 1.20\) & \(\mathbf{5.47 \pm 0.78}\)
            & \(4.38 \pm 1.18\) & \(4.48 \pm 1.24\) \\
        & Tool calls & \(48.1 \pm 9.1\) & \(\mathbf{18.1 \pm 5.3}\)
            & \(28.4 \pm 1.6\) & \(25.8 \pm 4.0\) \\
        \midrule
        \shortstack[l]{\texttt{075ef4df}\\{[}Hard, 14{]}}
            & Reward & \(2.96 \pm 2.37\) & \(4.43 \pm 1.56\)
            & \(4.84 \pm 0.34\) & \(\mathbf{5.09 \pm 1.86}\) \\
        & Tool calls & \(45.2 \pm 12.6\) & \(\mathbf{14.4 \pm 2.2}\)
            & \(18.4 \pm 2.7\) & \(17.7 \pm 1.3\) \\
        \bottomrule
    \end{tabular}
    \caption{Absolute APEX results underlying Table~\ref{tab:apex-results}:
    for each world [difficulty, \(N\)], Reward is total strict pass-discounted
    reward and Tool calls are benchmark-counted task-agent Archipelago
    calls per question excluding memory/harness calls; values are run means \(\pm\)
    95\% Student-\(t\) confidence intervals (\(n=3\) for GHCP (No Memory),
    \(n=5\) otherwise), with highest reward and lowest calls bolded.}
    \label{tab:apex-results-detailed}
\end{table}

\FloatBarrier

\section{Prompt Templates}
\label{app:prompt-templates}
\lstdefinestyle{memoryprompt}{
    numbers=none,
    xleftmargin=0pt,
    xrightmargin=0pt,
    framexleftmargin=0pt,
    framexrightmargin=0pt,
    linewidth=\linewidth,
    resetmargins=true,
    frame=single,
    columns=fullflexible,
    keepspaces=true,
    aboveskip=2pt,
    belowskip=6pt
}
\lstdefinestyle{probeaddition}{
    style=memoryprompt,
    backgroundcolor=\color{workedEnvProbeFrame!8!white},
    rulecolor=\color{workedEnvProbeFrame}
}

\subsection{Distiller Preprocessing Prompt}
\label{app:distiller-prompt}

The distiller is a pure preprocessing step enabled for both memory
conditions in our experiments.  It sees the completed raw trajectory, but not
terminal benchmark feedback, and has no memory or environment tools.  Its
output is a compact evidence packet for the curator agent; it cannot create, update,
or delete durable records.

\noindent\textbf{System message.}\par\nobreak
\begin{lstlisting}[style=memoryprompt]
ROLE
You are the non-writing trajectory distiller in a continual-learning
memory pipeline.

BOUNDARY
This is preprocessing only. You have no memory tools and must not
create, update, delete, or propose durable memory records. You do not
receive terminal benchmark feedback. Treat the supplied rollout as
partial evidence, not as ground truth.

INPUT
One completed task-agent trajectory containing the user request,
retrieved memories, assistant messages, tool calls, tool results,
environment observations, and submitted answer.

OBJECTIVE
Transform the raw trajectory into a compact, faithful evidence packet
that helps a later curator agent decide what is reusable. Do not evaluate
memory policy or add facts that are absent from the trajectory.

PRESERVE
- the task goal, constraints, and response requirements;
- retrieved memories and how the agent used or contradicted them;
- the chronological strategy and decisive action-observation pairs;
- successful and failed procedures, tool conventions, and environment
  structure discovered during execution;
- the submitted answer, unresolved questions, and assumptions that
  remain unverified.

OUTPUT
Return exactly these tagged sections:
<overview>task, constraints, and approach</overview>
<history>chronological actions and observations</history>
<work_done>completion state and submitted result</work_done>
<technical_details>reusable findings, failures, and quirks</technical_details>
<important_files>files or resources central to the task</important_files>
<next_steps>unresolved work and unverified assumptions</next_steps>
<checkpoint_title>a concise 2-6 word title</checkpoint_title>

Be concise, but retain evidence that would be costly to rediscover.
Refer to the task agent in the third person.
\end{lstlisting}

\noindent\textbf{Per-instance user message.}\par\nobreak
\begin{lstlisting}[style=memoryprompt]
Instance: {INSTANCE_ID}

Preprocess the completed task-agent rollout below.

<trajectory>
{RAW_TRAJECTORY}
</trajectory>
\end{lstlisting}

\subsection{Trajectory-Only Curator Agent Prompt}
\label{app:curator-prompt}

GHCP + Mem uses the following compact prompt for the curator agent.  Unlike
the distiller, this post-task curator agent receives terminal feedback and memory
CRUD tools.  It may inspect the staged raw trace when the distilled trajectory
omits a needed detail, but it has no live task-environment tools.

\noindent\textbf{System message.}\par\nobreak
\begin{lstlisting}[style=memoryprompt]
ROLE
You are the post-task curator agent for a continual-learning run.
Only you may mutate the persistent memory index.

INPUTS
1. A distilled evidence packet for one completed task-agent rollout.
2. Terminal feedback for that rollout, when available.
3. Relevant existing memory records.
4. An optional staged raw trajectory for evidence lookup.

GOAL
Maintain a small set of reliable, transferable, and actionable
memories that lets a future agent solve related tasks more accurately
and with fewer environment calls.

AVAILABLE TOOLS
- memory_read, memory_create, memory_update, memory_delete;
- sandbox readers for inspecting the staged raw trajectory.

RECORD CONTRACT
Every created record must contain:
- category: pattern | rule | trap | schema | policy | interaction;
- confidence: high | medium | low;
- applies_to: a short retrieval scope;
- lemma: one concise, actionable claim.

PROCEDURE
1. PROPOSE atomic candidate memories from the evidence packet,
   feedback, raw trace as needed, and nearby existing records.
2. CHECK each candidate for evidential support, transfer value, scope,
   actionability, current validity, and redundancy. A successful task
   does not validate every intermediate assumption.
3. RECONCILE with existing memory:
   - Create a supported, nonredundant candidate.
   - Update or merge when evidence refines an existing record.
   - Narrow, correct, or delete a contradicted record.
   - Skip unsupported, trivial, or instance-specific content.
4. COMMIT the minimum CRUD operations needed to leave a coherent index.

WRITING RULES
- Store procedures, relations, conventions, and scoped warnings; never
  memorize the task answer, rubric wording, or incidental values.
- Prefer positive rules that tell the next agent what to do.
- Scope no claim more broadly than its evidence supports.
- Prefer fewer, stronger records over many noisy ones.

STOP
When no further justified CRUD operation remains, stop using tools and
briefly summarize what changed.
\end{lstlisting}

\subsection{Environment-Probing Curator Agent Prompt}
\label{app:probing-curator-prompt}

The environment-probing system message deliberately preserves the curator
agent's core prompt above.  It is formed by inserting only two probe-specific
blocks:
\[
P_{\mathrm{probe}}
=P_{\mathrm{curator}}+A_{\mathrm{access}}+A_{\mathrm{verify}}.
\]
Reading the shared prompt for the curator agent with the two blue-highlighted blocks below
inserted at the stated locations gives the complete compact probing prompt.
No input, memory schema, CRUD policy, or stopping rule otherwise changes.

\noindent\textbf{Highlighted addition 1: read-only tool access.}\par\nobreak
\noindent Insert under \textsc{Available Tools}.\par\nobreak
\begin{lstlisting}[style=probeaddition]
ENVIRONMENT-PROBING ADDITION
You also have read-only task-environment tools. These tools cannot
mutate the environment, consume the task agent's budget, or expose
future tasks or labels.
\end{lstlisting}

\noindent\textbf{Highlighted addition 2: verification nudge.}\par\nobreak
\noindent Insert between \textsc{Check} and \textsc{Reconcile}.\par\nobreak
\begin{lstlisting}[style=probeaddition]
ENVIRONMENT-PROBING ADDITION
Use the read-only environment tools to verify candidate and existing
memories before Create or Update whenever correctness, scope,
freshness, or actionability is uncertain. Probe counterexamples,
untouched slices, stale mappings, required preconditions, and whether
a shorter procedure yields the same evidence.

Probe only to evaluate a candidate memory, not to solve a future task
or explore without a hypothesis. Use probe evidence to strengthen or
narrow a supported record and to update or delete a contradicted one.
\end{lstlisting}

\subsection{Shared Curator Agent User-Message Template}
\label{app:curator-user-message}

Both curation conditions receive the same dynamic evidence message.  The raw
trajectory pointer is omitted only when no staged trace is available.

\begin{lstlisting}[style=memoryprompt]
<distilled_evidence>
{DISTILLED_TRAJECTORY}
</distilled_evidence>

<terminal_feedback>
{TERMINAL_FEEDBACK_OR_NONE}
</terminal_feedback>

<related_memory>
{RELATED_MEMORY_ENTRIES_OR_NONE}
</related_memory>

<raw_trajectory path="{TRAJECTORY_JSON_PATH_OR_NONE}" />

Reconcile the memory index using the available tools. Stop when no
further justified operation remains.
\end{lstlisting}

\section{Memory Samples and Trajectory Side-by-Sides}
\label{app:worked-examples}

This appendix first compares representative curator-agent memory records from
the CLBench database-exploration runs.  It then expands two matched cases, one
from each benchmark, as trajectory side-by-sides.

\subsection{CLBench Curator Memory Examples}
\label{app:clbench-memory-record-examples}
Figure~\ref{fig:clbench-curator-memory-records} shows examples from the
CLBench database-exploration task runs and the kinds of memories generated by
the curator agent.  The left column samples trajectory-only curation; the
right column samples the same memory schema when the curator agent can make
targeted read-only environment probes.  Record text is lightly shortened for
layout, while table names, fields, relations, and values are preserved.

\begin{center}
\centering
\begin{minipage}[t]{0.485\linewidth}
\centering
{\large\sffamily\bfseries\color{workedMemoryFrame} GHCP + Mem\par}
{\small\sffamily Trajectory-only curator agent\par}
\smallskip

\begin{tcolorbox}[
    colback=workedMemoryFrame!6!white,
    colframe=workedMemoryFrame,
    title={\texttt{[trap]} Answer-Anchored Warning},
    fonttitle=\small\sffamily\bfseries,
    fontupper=\footnotesize,
    boxrule=0.65pt,
    arc=1mm,
    left=4pt,right=4pt,top=3pt,bottom=3pt,
    before upper={\raggedright}
]
\textbf{\texttt{applies\_to:}} electronics average-listed-price questions
\par\smallskip
\textbf{\texttt{lemma:}} Do not answer with
\texttt{AVG(items\_g2.prc\_usd)} over non-null rows; that produces about
52.96, but the benchmark's correct result is 96.23, so a different price
field and/or row subset is required.
\end{tcolorbox}

\begin{tcolorbox}[
    colback=workedMemoryFrame!6!white,
    colframe=workedMemoryFrame,
    title={\texttt{[schema]} Broad Map, Missing Relation},
    fonttitle=\small\sffamily\bfseries,
    fontupper=\footnotesize,
    boxrule=0.65pt,
    arc=1mm,
    left=4pt,right=4pt,top=3pt,bottom=3pt,
    before upper={\raggedright}
]
\textbf{\texttt{applies\_to:}} grouped product and review tables
\par\smallskip
\textbf{\texttt{lemma:}} Across the grouped product/review tables,
\texttt{g1} = office products, \texttt{g2} = electronics, and \texttt{g3} =
musical instruments; this mapping applies to both \texttt{items\_g*} and
\texttt{fdbk\_g*}.
\end{tcolorbox}

\begin{tcolorbox}[
    colback=workedMemoryFrame!6!white,
    colframe=workedMemoryFrame,
    title={\texttt{[rule]} Stale Table Name},
    fonttitle=\small\sffamily\bfseries,
    fontupper=\footnotesize,
    boxrule=0.65pt,
    arc=1mm,
    left=4pt,right=4pt,top=3pt,bottom=3pt,
    before upper={\raggedright}
]
\textbf{\texttt{applies\_to:}} musical-instrument brand questions
\par\smallskip
\textbf{\texttt{lemma:}} Use \texttt{attrs\_g3} (not
\texttt{items\_g3}); filter \texttt{attr\_key='Brand'}, ignore blank
\texttt{attr\_val}, and count distinct \texttt{ref\_id}.
\par\smallskip
\emph{After migration, \texttt{attrs\_g3} no longer exists.}
\end{tcolorbox}
\end{minipage}
\hfill
\begin{minipage}[t]{0.485\linewidth}
\centering
{\large\sffamily\bfseries\color{workedEnvProbeFrame}
GHCP + Mem (w/ Env Probing)\par}
{\small\sffamily Environment-probing curator agent\par}
\smallskip

\begin{tcolorbox}[
    colback=workedEnvProbeFrame!6!white,
    colframe=workedEnvProbeFrame,
    title={\texttt{[rule]} Executable Join and Filter},
    fonttitle=\small\sffamily\bfseries,
    fontupper=\footnotesize,
    boxrule=0.65pt,
    arc=1mm,
    left=4pt,right=4pt,top=3pt,bottom=3pt,
    before upper={\raggedright}
]
\textbf{\texttt{applies\_to:}} \texttt{g2} top-level-category price shares
\par\smallskip
\textbf{\texttt{lemma:}} Join \texttt{items\_g2} to \texttt{taxn\_g2} on
\texttt{ref\_id}; filter \texttt{cat\_lvl=1} and the exact
\texttt{cat\_nm}; keep \texttt{items\_g2.prc>0}; then compare against the
filtered \texttt{AVG(prc)}.
\end{tcolorbox}

\begin{tcolorbox}[
    colback=workedEnvProbeFrame!6!white,
    colframe=workedEnvProbeFrame,
    title={\texttt{[rule]} Positive Relation and Grain},
    fonttitle=\small\sffamily\bfseries,
    fontupper=\footnotesize,
    boxrule=0.65pt,
    arc=1mm,
    left=4pt,right=4pt,top=3pt,bottom=3pt,
    before upper={\raggedright}
]
\textbf{\texttt{applies\_to:}} grouped review-average questions
\par\smallskip
\textbf{\texttt{lemma:}} Use \texttt{items\_g*} for product-side filters,
join to \texttt{fdbk\_g*} on \texttt{ref\_id}, and aggregate
\texttt{fdbk\_g*.rtg} rather than averaging item-side
\texttt{avg\_rtg}.
\end{tcolorbox}

\begin{tcolorbox}[
    colback=workedEnvProbeFrame!6!white,
    colframe=workedEnvProbeFrame,
    title={\texttt{[schema]} Current Migrated Schema},
    fonttitle=\small\sffamily\bfseries,
    fontupper=\footnotesize,
    boxrule=0.65pt,
    arc=1mm,
    left=4pt,right=4pt,top=3pt,bottom=3pt,
    before upper={\raggedright}
]
\textbf{\texttt{applies\_to:}} musical-instrument attributes and brands
\par\smallskip
\textbf{\texttt{lemma:}} Use \texttt{product\_attributes\_g3} for musical
attributes.  For brand questions, filter \texttt{attr\_key='Brand'} with
nonblank \texttt{attr\_val}, group by \texttt{attr\_val}, and count distinct
\texttt{ref\_id}.
\end{tcolorbox}
\end{minipage}
\captionof{figure}{Representative CLBench memory records generated by the curator
agent.  Trajectory-only curation (left) can preserve a rejected answer,
underspecify the positive operation, or retain a stale schema name.
Environment probing (right) yields positive procedures with explicit joins,
filters, aggregation grain, and current schema.}
\label{fig:clbench-curator-memory-records}
\end{center}

The trajectory-only records are not uniformly wrong: they recover useful
domain mappings and warnings.  Their weakness is actionability.  The first
record says what failed but not what should replace it; the second orients the
agent to the domain but does not identify the relation needed by the current
question; and the third is a once-valid rule that survived schema drift.  A
later task agent therefore has to reconstruct the missing evidence.

The probe-backed records instead verbalize observations that can be executed
directly: which tables to use, how they join, which rows define the
denominator, what aggregation grain is valid, and which schema name is
current.  In the three matched CLBench examples, GHCP + Mem used 4, 9, and 8
queries, whereas GHCP + Mem (w/ Env Probing) used 1, 2, and 1.  The examples
do not imply that every probed record is complete, but they show how read-only
checks can convert a warning or tentative mapping into an environment-informed procedure
before a future task retrieves it.

The side-by-sides hold the task fixed across \textsc{Base} (GHCP (No
Memory)), \textsc{Mem} (GHCP + Mem), and \textsc{Probe} (GHCP + Mem (w/ Env
Probing)).  The lanes show task-time behavior---memory retrieval, selected
tool calls, the final answer, and grader evidence---not the earlier
post-task sessions that produced the retrieved records.

For both memory conditions in the reported runs, a non-writing
distiller transformed prior completed trajectories into evidence packets,
after which a separate curator agent used memory CRUD to reconcile candidate lemmas
with the persistent index.
Only \textsc{Probe} gave that curator agent the read-only benchmark tools and
the highlighted verification instructions in
Appendix~\ref{app:prompt-templates}.  During the displayed task, both memory
conditions expose only \texttt{memory\_read}, and all three responding agents
retain the same environment-tool surface.  The comparison therefore isolates
the task-time effect of previously curated memory, including the additional
environment evidence supplied by curator-side probes.

Each case appears as sequential red (\textsc{Base}), gold
(\textsc{Mem}), and blue (\textsc{Probe}) lanes.  Agent calls and submitted
answers are right-aligned; memory results, environment observations, call
summaries, and grader evidence are left-aligned.  Lane headers report
task-agent response cost and input, output, and cached tokens; distiller and
curator-agent usage are not included.

These are selected-evidence traces rather than full transcripts.  CLBench
shows the first and final SQL queries; APEX shows the first two discovery calls
and final computation call.  Each displayed call gives its position in the
recorded trace, and bucket cards account for every omitted call without
printing its arguments or result.  APEX positions count recorded top-level
calls, while lane headers report total task-agent tool calls.
Tagged observations are capped at 500 source characters and carry an
original-length marker when truncated.  Memory payloads, final answers, and
grader evidence are reproduced from the recorded runs; internal model
reasoning is omitted.

\subsection{CLBench Database Exploration}
\label{app:worked-clbench}

The selected CLBench case captures both correctness and efficiency on a
question about products that have reviews but no attributes row.  The
stateless agent adds an incorrect category condition and answers 188 after
seven database queries.  GHCP + Mem retrieves useful mappings and
warnings but not the required positive join pattern; it explores for nine
queries and answers 267.  GHCP + Mem (w/ Env Probing) retrieves the curator
agent's
live-validated \texttt{ref\_id} relation and answers 267 in two queries.  A
wrong answer receives zero reward under
Equation~\ref{eq:main-pass-discounted-reward}.

\subsubsection{C2: Filling a Knowledge Blind Spot.}
\label{app:worked-clbench-c2}
\noindent\textit{Run 3;\allowbreak{} position \#7 of 40;\allowbreak{} memory conditions retrieve records curated from prior completed tasks.\allowbreak{}}\par

\begin{workedshared}{workedAgentBg}{QUESTION AND GROUND TRUTH}
{\sffamily\bfseries Question}\par
How many office products have at least one review in the database but are NOT represented in the attributes data at all?\par
\smallskip
{\sffamily\bfseries Ground truth}\par
Accepted answer:\allowbreak{} 267.\allowbreak{}\par
\end{workedshared}

\begin{workedlaneheader}{workedFailFrame}{GHCP --- NO MEMORY}
config:\allowbreak{} GHCP (No Memory)\allowbreak{}\par
status:\allowbreak{} failure | reward:\allowbreak{} 0.\allowbreak{}00 | queries:\allowbreak{} 7 | 8 turns\par
cost:\allowbreak{} \$0.\allowbreak{}0684 | in:\allowbreak{} 63.\allowbreak{}1K | out:\allowbreak{} 1.\allowbreak{}5K | cached:\allowbreak{} 49.\allowbreak{}9K\par
\end{workedlaneheader}
\begin{workedlanesection}{workedFailFrame}
MEMORY
\end{workedlanesection}
\begin{workedsystembubble}{workedResultBg}{workedFailFrame}{memory retrieval}
No memory is available to the stateless baseline.\allowbreak{}\par
\end{workedsystembubble}
\begin{workedlanesection}{workedFailFrame}
SELECTED AGENT CALLS
\end{workedlanesection}
\begin{workedagentbubble}{workedFailFrame}{agent \textperiodcentered{} query \textperiodcentered{} representative discovery}
Recorded query 1 of 7 | 0 before | 6 after\par
\smallskip
.\allowbreak{}tables\par
\end{workedagentbubble}
\begin{workedsystembubble}{workedResultBg}{workedFailFrame}{env \textperiodcentered{} observation \textperiodcentered{} representative discovery}
Query result (1/\allowbreak{}15 queries used,\allowbreak{} 14 remaining)\allowbreak{}:\allowbreak{}\par
\smallskip
attrs\_\allowbreak{}g1\par
attrs\_\allowbreak{}g3\par
fdbk\_\allowbreak{}g1\par
fdbk\_\allowbreak{}g2\par
fdbk\_\allowbreak{}g3\par
fdbk\_\allowbreak{}stats\_\allowbreak{}g1\par
items\_\allowbreak{}g1\par
items\_\allowbreak{}g2\par
items\_\allowbreak{}g3\par
sqlite\_\allowbreak{}sequence\par
taxn\_\allowbreak{}g1\par
taxn\_\allowbreak{}g2\par
\end{workedsystembubble}
\begin{workedagentbubble}{workedFailFrame}{agent \textperiodcentered{} query \textperiodcentered{} decisive computation}
Recorded query 7 of 7 | 6 before | 0 after\par
\smallskip
SELECT COUNT(*)\allowbreak{} AS cnt FROM items\_\allowbreak{}g1 i WHERE i.\allowbreak{}main\_\allowbreak{}cat='Office Products' AND EXISTS (SELECT 1 FROM fdbk\_\allowbreak{}g1 f WHERE f.\allowbreak{}ref\_\allowbreak{}id=i.\allowbreak{}ref\_\allowbreak{}id)\allowbreak{} AND NOT EXISTS (SELECT 1 FROM attrs\_\allowbreak{}g1 a WHERE a.\allowbreak{}ref\_\allowbreak{}id=i.\allowbreak{}ref\_\allowbreak{}id)\allowbreak{};\allowbreak{}\par
\end{workedagentbubble}
\begin{workedsystembubble}{workedResultBg}{workedFailFrame}{env \textperiodcentered{} observation \textperiodcentered{} decisive computation}
Query result (7/\allowbreak{}15 queries used,\allowbreak{} 8 remaining)\allowbreak{}:\allowbreak{}\par
\smallskip
cnt\par
-\allowbreak{}-\allowbreak{}-\allowbreak{}\par
188\par
\end{workedsystembubble}
\begin{workedlanesection}{workedFailFrame}
OTHER RECORDED CALLS BY BUCKET
\end{workedlanesection}
\begin{workedsystembubble}{workedResultBg}{workedFailFrame}{other recorded queries}
Other recorded queries (not expanded)\allowbreak{}:\allowbreak{} schema inspection x3;\allowbreak{} data exploration/\allowbreak{}validation x2.\allowbreak{} Accounting:\allowbreak{} 7 recorded;\allowbreak{} 2 shown;\allowbreak{} 5 not expanded.\allowbreak{}\par
\end{workedsystembubble}
\begin{workedlanesection}{workedFailFrame}
FINAL ANSWER AND GRADING
\end{workedlanesection}
\begin{workedagentbubble}{workedFailFrame}{submitted answer}
188\par
\end{workedagentbubble}
\begin{workedsystembubble}{workedFailBg}{workedFailFrame}{grader evidence \textperiodcentered{} FAIL}
Question 1:\allowbreak{} INCORRECT.\allowbreak{}\par
Your answer:\allowbreak{} 188\par
Correct answer:\allowbreak{} 267\par
Exploratory queries used:\allowbreak{} 7\par
\end{workedsystembubble}
\par\vspace{6pt}
\begin{workedlaneheader}{workedMemoryFrame}{GHCP + Mem}
config:\allowbreak{} GHCP + Mem\par
status:\allowbreak{} success | reward:\allowbreak{} 0.\allowbreak{}40 | queries:\allowbreak{} 9 | 10 turns\par
cost:\allowbreak{} \$0.\allowbreak{}0983 | in:\allowbreak{} 94.\allowbreak{}7K | out:\allowbreak{} 2.\allowbreak{}7K | cached:\allowbreak{} 79.\allowbreak{}5K\par
\end{workedlaneheader}
\begin{workedlanesection}{workedMemoryFrame}
MEMORY
\end{workedlanesection}
\begin{workedagentbubble}{workedMemoryFrame}{agent \textperiodcentered{} memory\_\allowbreak{}read}
\{\par
  "query":\allowbreak{} "Question 7:\allowbreak{} determine count of office products with at least one review but no attributes representation;\allowbreak{} need identify relevant tables and likely join keys.\allowbreak{}",\allowbreak{}\par
  "k":\allowbreak{} 5\par
\}\allowbreak{}\par
\end{workedagentbubble}
\begin{workedsystembubble}{workedResultBg}{workedMemoryFrame}{memory \textperiodcentered{} complete logical payload}
\#1 [trap]\allowbreak{}\par
applies\_\allowbreak{}to:\allowbreak{} SQLite product-\allowbreak{}dataset benchmark questions about reviewers spanning 2+ top-\allowbreak{}level categories within office products\par
lemma:\allowbreak{} Do not treat grouped datasets g1/\allowbreak{}g2/\allowbreak{}g3 as the requested 'top-\allowbreak{}level categories';\allowbreak{} for office-\allowbreak{}products category-\allowbreak{}span questions,\allowbreak{} join office reviews (fdbk\_\allowbreak{}g1)\allowbreak{} to office item/\allowbreak{}taxonomy data (e.\allowbreak{}g.\allowbreak{} items\_\allowbreak{}g1/\allowbreak{}taxn\_\allowbreak{}*)\allowbreak{} and count distinct top-\allowbreak{}level categories within the office domain.\allowbreak{}\par
\smallskip
\#2 [schema]\allowbreak{}\par
applies\_\allowbreak{}to:\allowbreak{} SQLite product-\allowbreak{}dataset benchmark with grouped items\_\allowbreak{}g* and fdbk\_\allowbreak{}g* tables\par
lemma:\allowbreak{} Across the grouped product/\allowbreak{}review tables,\allowbreak{} g1 = office products,\allowbreak{} g2 = electronics,\allowbreak{} and g3 = musical instruments;\allowbreak{} this mapping applies to both items\_\allowbreak{}g* and fdbk\_\allowbreak{}g* tables.\allowbreak{}\par
\smallskip
\#3 [schema]\allowbreak{}\par
applies\_\allowbreak{}to:\allowbreak{} SQLite product-\allowbreak{}dataset benchmark questions over grouped feedback tables fdbk\_\allowbreak{}g* that filter verified or non-\allowbreak{}empty review text\par
lemma:\allowbreak{} In grouped feedback tables fdbk\_\allowbreak{}g*,\allowbreak{} rtg is the review rating,\allowbreak{} body is the review text,\allowbreak{} vrf is the verification flag (e.\allowbreak{}g.\allowbreak{} filter verified with vrf = 'true')\allowbreak{},\allowbreak{} and non-\allowbreak{}empty text should be filtered with body IS NOT NULL AND TRIM(body)\allowbreak{} \textless{}\textgreater{} ''.\allowbreak{}\par
\smallskip
\#4 [trap]\allowbreak{}\par
applies\_\allowbreak{}to:\allowbreak{} SQLite product-\allowbreak{}dataset benchmark questions asking for average review rating by an item attribute (for example,\allowbreak{} whether price is listed)\allowbreak{}\par
lemma:\allowbreak{} Do not answer 'average review rating' questions with AVG(items\_\allowbreak{}*.\allowbreak{}avg\_\allowbreak{}rtg)\allowbreak{};\allowbreak{} the benchmark expects averaging review rows from fdbk\_\allowbreak{}*.\allowbreak{}rtg after joining to items\_\allowbreak{}* only to classify products by the requested attribute.\allowbreak{}\par
\smallskip
\#5 [trap]\allowbreak{}\par
applies\_\allowbreak{}to:\allowbreak{} Questions about average listed price for electronics in the grouped SQLite product datasets\par
lemma:\allowbreak{} Do not answer electronics 'average listed price' with AVG(items\_\allowbreak{}g2.\allowbreak{}prc\_\allowbreak{}usd)\allowbreak{} over non-\allowbreak{}null rows;\allowbreak{} that produces about 52.\allowbreak{}96,\allowbreak{} but the benchmark's correct result is 96.\allowbreak{}23,\allowbreak{} so a different price field and/\allowbreak{}or row subset is required.\allowbreak{}\par
\end{workedsystembubble}
\begin{workedlanesection}{workedMemoryFrame}
SELECTED AGENT CALLS
\end{workedlanesection}
\begin{workedagentbubble}{workedMemoryFrame}{agent \textperiodcentered{} query \textperiodcentered{} representative discovery}
Recorded query 1 of 9 | 0 before | 8 after\par
\smallskip
.\allowbreak{}tables\par
\end{workedagentbubble}
\begin{workedsystembubble}{workedResultBg}{workedMemoryFrame}{env \textperiodcentered{} observation \textperiodcentered{} representative discovery}
Query result (1/\allowbreak{}15 queries used,\allowbreak{} 14 remaining)\allowbreak{}:\allowbreak{}\par
\smallskip
attrs\_\allowbreak{}g1\par
attrs\_\allowbreak{}g3\par
fdbk\_\allowbreak{}g1\par
fdbk\_\allowbreak{}g2\par
fdbk\_\allowbreak{}g3\par
fdbk\_\allowbreak{}stats\_\allowbreak{}g1\par
items\_\allowbreak{}g1\par
items\_\allowbreak{}g2\par
items\_\allowbreak{}g3\par
sqlite\_\allowbreak{}sequence\par
taxn\_\allowbreak{}g1\par
taxn\_\allowbreak{}g2\par
\end{workedsystembubble}
\begin{workedagentbubble}{workedMemoryFrame}{agent \textperiodcentered{} query \textperiodcentered{} decisive computation}
Recorded query 9 of 9 | 8 before | 0 after\par
\smallskip
SELECT (SELECT COUNT(*)\allowbreak{} FROM (SELECT DISTINCT ref\_\allowbreak{}id FROM fdbk\_\allowbreak{}g1 WHERE ref\_\allowbreak{}id IS NOT NULL EXCEPT SELECT ref\_\allowbreak{}id FROM items\_\allowbreak{}g1)\allowbreak{})\allowbreak{} AS review\_\allowbreak{}refids\_\allowbreak{}not\_\allowbreak{}in\_\allowbreak{}items,\allowbreak{} (SELECT COUNT(*)\allowbreak{} FROM (SELECT DISTINCT ref\_\allowbreak{}id FROM items\_\allowbreak{}g1 WHERE ref\_\allowbreak{}id IN (SELECT ref\_\allowbreak{}id FROM fdbk\_\allowbreak{}g1 WHERE ref\_\allowbreak{}id IS NOT NULL)\allowbreak{} EXCEPT SELECT DISTINCT ref\_\allowbreak{}id FROM attrs\_\allowbreak{}g1 WHERE ref\_\allowbreak{}id IS NOT NULL)\allowbreak{})\allowbreak{} AS item\_\allowbreak{}refids\_\allowbreak{}with\_\allowbreak{}review\_\allowbreak{}missing\_\allowbreak{}attrs\par
\end{workedagentbubble}
\begin{workedsystembubble}{workedResultBg}{workedMemoryFrame}{env \textperiodcentered{} observation \textperiodcentered{} decisive computation}
Query result (9/\allowbreak{}15 queries used,\allowbreak{} 6 remaining)\allowbreak{}:\allowbreak{}\par
\smallskip
review\_\allowbreak{}refids\_\allowbreak{}not\_\allowbreak{}in\_\allowbreak{}items | item\_\allowbreak{}refids\_\allowbreak{}with\_\allowbreak{}review\_\allowbreak{}missing\_\allowbreak{}attrs\par
-\allowbreak{}-\allowbreak{}-\allowbreak{}-\allowbreak{}-\allowbreak{}-\allowbreak{}-\allowbreak{}-\allowbreak{}-\allowbreak{}-\allowbreak{}-\allowbreak{}-\allowbreak{}-\allowbreak{}-\allowbreak{}-\allowbreak{}-\allowbreak{}-\allowbreak{}-\allowbreak{}-\allowbreak{}-\allowbreak{}-\allowbreak{}-\allowbreak{}-\allowbreak{}-\allowbreak{}-\allowbreak{}-\allowbreak{}-\allowbreak{}+-\allowbreak{}-\allowbreak{}-\allowbreak{}-\allowbreak{}-\allowbreak{}-\allowbreak{}-\allowbreak{}-\allowbreak{}-\allowbreak{}-\allowbreak{}-\allowbreak{}-\allowbreak{}-\allowbreak{}-\allowbreak{}-\allowbreak{}-\allowbreak{}-\allowbreak{}-\allowbreak{}-\allowbreak{}-\allowbreak{}-\allowbreak{}-\allowbreak{}-\allowbreak{}-\allowbreak{}-\allowbreak{}-\allowbreak{}-\allowbreak{}-\allowbreak{}-\allowbreak{}-\allowbreak{}-\allowbreak{}-\allowbreak{}-\allowbreak{}-\allowbreak{}-\allowbreak{}-\allowbreak{}-\allowbreak{}-\allowbreak{}\par
0                          | 267                                  \par
\end{workedsystembubble}
\begin{workedlanesection}{workedMemoryFrame}
OTHER RECORDED CALLS BY BUCKET
\end{workedlanesection}
\begin{workedsystembubble}{workedResultBg}{workedMemoryFrame}{other recorded queries}
Other recorded queries (not expanded)\allowbreak{}:\allowbreak{} schema inspection x4;\allowbreak{} data exploration/\allowbreak{}validation x3.\allowbreak{} Accounting:\allowbreak{} 9 recorded;\allowbreak{} 2 shown;\allowbreak{} 7 not expanded.\allowbreak{}\par
\end{workedsystembubble}
\begin{workedlanesection}{workedMemoryFrame}
FINAL ANSWER AND GRADING
\end{workedlanesection}
\begin{workedagentbubble}{workedMemoryFrame}{submitted answer}
267\par
\end{workedagentbubble}
\begin{workedsystembubble}{workedPassBg}{workedMemoryFrame}{grader evidence \textperiodcentered{} PASS}
Question 7:\allowbreak{} CORRECT!\par
Your answer:\allowbreak{} 267\par
Exploratory queries used:\allowbreak{} 9\par
\end{workedsystembubble}
\par\vspace{6pt}
\begin{workedlaneheader}{workedEnvProbeFrame}{GHCP + Mem (w/\allowbreak{} Env Probing)\allowbreak{}}
config:\allowbreak{} GHCP + Mem (w/\allowbreak{} Env Probing)\allowbreak{}\par
status:\allowbreak{} success | reward:\allowbreak{} 0.\allowbreak{}87 | queries:\allowbreak{} 2 | 3 turns\par
cost:\allowbreak{} \$0.\allowbreak{}0214 | in:\allowbreak{} 25.\allowbreak{}0K | out:\allowbreak{} 568 | cached:\allowbreak{} 22.\allowbreak{}0K\par
\end{workedlaneheader}
\begin{workedlanesection}{workedEnvProbeFrame}
MEMORY
\end{workedlanesection}
\begin{workedagentbubble}{workedEnvProbeFrame}{agent \textperiodcentered{} memory\_\allowbreak{}read}
\{\par
  "query":\allowbreak{} "Question 7:\allowbreak{} count office products with at least one review but no attributes rows;\allowbreak{} need identify office tables and likely join key",\allowbreak{}\par
  "k":\allowbreak{} 5\par
\}\allowbreak{}\par
\end{workedagentbubble}
\begin{workedsystembubble}{workedResultBg}{workedEnvProbeFrame}{memory \textperiodcentered{} complete logical payload}
\#1 [rule]\allowbreak{}\par
applies\_\allowbreak{}to:\allowbreak{} g1 reviewer category-\allowbreak{}span queries\par
lemma:\allowbreak{} For office-\allowbreak{}products reviewer span questions,\allowbreak{} join fdbk\_\allowbreak{}g1 to taxn\_\allowbreak{}g1 by ref\_\allowbreak{}id,\allowbreak{} treat taxn\_\allowbreak{}g1.\allowbreak{}cat\_\allowbreak{}lvl=1 cat\_\allowbreak{}nm as the top-\allowbreak{}level category,\allowbreak{} count reviewers with at least 2 distinct cat\_\allowbreak{}nm,\allowbreak{} and divide by COUNT(DISTINCT uid)\allowbreak{} from fdbk\_\allowbreak{}g1.\allowbreak{}\par
\smallskip
\#2 [rule]\allowbreak{}\par
applies\_\allowbreak{}to:\allowbreak{} g1/\allowbreak{}g3 brand product counts\par
lemma:\allowbreak{} For office-\allowbreak{}products or musical-\allowbreak{}instruments brand questions,\allowbreak{} read brand from attrs\_\allowbreak{}g1/\allowbreak{}attrs\_\allowbreak{}g3 by filtering attr\_\allowbreak{}key='Brand' and aggregate COUNT(DISTINCT ref\_\allowbreak{}id)\allowbreak{} by attr\_\allowbreak{}val.\allowbreak{}\par
\smallskip
\#3 [rule]\allowbreak{}\par
applies\_\allowbreak{}to:\allowbreak{} grouped average review rating queries\par
lemma:\allowbreak{} For g1/\allowbreak{}g2/\allowbreak{}g3 questions comparing average review rating across product attributes,\allowbreak{} join fdbk\_\allowbreak{}g* to items\_\allowbreak{}g* on ref\_\allowbreak{}id and average fdbk\_\allowbreak{}g*.\allowbreak{}rtg within each attribute-\allowbreak{}defined group.\allowbreak{}\par
\smallskip
\#4 [rule]\allowbreak{}\par
applies\_\allowbreak{}to:\allowbreak{} g2 electronics review body-\allowbreak{}length queries\par
lemma:\allowbreak{} For electronics review-\allowbreak{}text metrics,\allowbreak{} aggregate directly from fdbk\_\allowbreak{}g2:\allowbreak{} use rtg for stars and body for text,\allowbreak{} filter verified reviews with vrf='true',\allowbreak{} exclude blank reviews with body IS NOT NULL AND TRIM(body)\allowbreak{}\textless{}\textgreater{}'' and compute measures like AVG(LENGTH(body)\allowbreak{})\allowbreak{} over the requested rating groups.\allowbreak{}\par
\smallskip
\#5 [schema]\allowbreak{}\par
applies\_\allowbreak{}to:\allowbreak{} grouped product domain mapping\par
lemma:\allowbreak{} In the g1/\allowbreak{}g2/\allowbreak{}g3 product split,\allowbreak{} use g1 for office products,\allowbreak{} g2 for electronics,\allowbreak{} and g3 for musical instruments.\allowbreak{}\par
\end{workedsystembubble}
\begin{workedlanesection}{workedEnvProbeFrame}
SELECTED AGENT CALLS
\end{workedlanesection}
\begin{workedagentbubble}{workedEnvProbeFrame}{agent \textperiodcentered{} query \textperiodcentered{} representative discovery}
Recorded query 1 of 2 | 0 before | 1 after\par
\smallskip
SELECT name FROM sqlite\_\allowbreak{}master WHERE type='table' ORDER BY name;\allowbreak{}\par
\end{workedagentbubble}
\begin{workedsystembubble}{workedResultBg}{workedEnvProbeFrame}{env \textperiodcentered{} observation \textperiodcentered{} representative discovery}
Query result (1/\allowbreak{}15 queries used,\allowbreak{} 14 remaining)\allowbreak{}:\allowbreak{}\par
\smallskip
name           \par
-\allowbreak{}-\allowbreak{}-\allowbreak{}-\allowbreak{}-\allowbreak{}-\allowbreak{}-\allowbreak{}-\allowbreak{}-\allowbreak{}-\allowbreak{}-\allowbreak{}-\allowbreak{}-\allowbreak{}-\allowbreak{}-\allowbreak{}\par
attrs\_\allowbreak{}g1       \par
attrs\_\allowbreak{}g3       \par
fdbk\_\allowbreak{}g1        \par
fdbk\_\allowbreak{}g2        \par
fdbk\_\allowbreak{}g3        \par
fdbk\_\allowbreak{}stats\_\allowbreak{}g1  \par
items\_\allowbreak{}g1       \par
items\_\allowbreak{}g2       \par
items\_\allowbreak{}g3       \par
sqlite\_\allowbreak{}sequence\par
taxn\_\allowbreak{}g1        \par
taxn\_\allowbreak{}g2        \par
\end{workedsystembubble}
\begin{workedagentbubble}{workedEnvProbeFrame}{agent \textperiodcentered{} query \textperiodcentered{} decisive computation}
Recorded query 2 of 2 | 1 before | 0 after\par
\smallskip
SELECT COUNT(DISTINCT f.\allowbreak{}ref\_\allowbreak{}id)\allowbreak{} AS missing\_\allowbreak{}attrs\_\allowbreak{}products FROM fdbk\_\allowbreak{}g1 f LEFT JOIN attrs\_\allowbreak{}g1 a ON f.\allowbreak{}ref\_\allowbreak{}id = a.\allowbreak{}ref\_\allowbreak{}id WHERE a.\allowbreak{}ref\_\allowbreak{}id IS NULL;\allowbreak{}\par
\end{workedagentbubble}
\begin{workedsystembubble}{workedResultBg}{workedEnvProbeFrame}{env \textperiodcentered{} observation \textperiodcentered{} decisive computation}
Query result (2/\allowbreak{}15 queries used,\allowbreak{} 13 remaining)\allowbreak{}:\allowbreak{}\par
\smallskip
missing\_\allowbreak{}attrs\_\allowbreak{}products\par
-\allowbreak{}-\allowbreak{}-\allowbreak{}-\allowbreak{}-\allowbreak{}-\allowbreak{}-\allowbreak{}-\allowbreak{}-\allowbreak{}-\allowbreak{}-\allowbreak{}-\allowbreak{}-\allowbreak{}-\allowbreak{}-\allowbreak{}-\allowbreak{}-\allowbreak{}-\allowbreak{}-\allowbreak{}-\allowbreak{}-\allowbreak{}-\allowbreak{}\par
267                   \par
\end{workedsystembubble}
\begin{workedlanesection}{workedEnvProbeFrame}
FINAL ANSWER AND GRADING
\end{workedlanesection}
\begin{workedagentbubble}{workedEnvProbeFrame}{submitted answer}
267\par
\end{workedagentbubble}
\begin{workedsystembubble}{workedPassBg}{workedEnvProbeFrame}{grader evidence \textperiodcentered{} PASS}
Question 7:\allowbreak{} CORRECT!\par
Your answer:\allowbreak{} 267\par
Exploratory queries used:\allowbreak{} 2\par
\end{workedsystembubble}

\subsection{Adapted APEX Management Consulting}
\label{app:worked-apex}

The selected APEX case asks which Impact site has the largest absolute
\(z\)-score for 2024 average monthly US revenue per head.  The stateless
agent exhausts 96 calls and returns the incorrect answer
\texttt{Darcylis}, \(z=1.29\), satisfying neither criterion.  GHCP + Mem
transfers the computation recipe and returns \texttt{Lorexa},
\(z=-1.60\), in 11 calls; GHCP + Mem (w/ Env Probing) returns the same correct
answer in six calls using its validated workbook map.  Consistent with
Equation~\ref{eq:main-pass-discounted-reward}, the lane headers report the
primary reward \(r_i\), fractional-criteria diagnostic
\(r_i^{\mathrm{frac}}\), and task-agent tool-call count \(q_i\).

\subsubsection{A3: Transferring a Computation Recipe.}
\label{app:worked-apex-a3}
\noindent\textit{Run 4;\allowbreak{} position \#11 of 15;\allowbreak{} memory conditions retrieve records curated from prior completed tasks.\allowbreak{}}\par

\begin{workedshared}{workedAgentBg}{QUESTION AND GROUND TRUTH}
{\sffamily\bfseries Question}\par
Can you please calculate the z score of US 2024 Average Monthly Revenue per Head,\allowbreak{} for all of Impact's sites?\par
Use a distribution of US 2024 Average Monthly Revenue per Head per site for all the sites in the attached file,\allowbreak{} which has the monthly US operational data for all of Impact's and competitor's sites.\allowbreak{} You can allocate the yearly revenue from the respective P\&L equally across all the respective sites and months.\allowbreak{}\par
\smallskip
Tell me here the z score of only the Impact site with the highest absolute z score,\allowbreak{} and the SiteID of this Impact site.\allowbreak{} Use the standard deviation formula for sample,\allowbreak{} not population.\allowbreak{} Round the final answer to two decimal places.\allowbreak{} Write back to me with what I've asked for.\allowbreak{}\par
\smallskip
{\sffamily\bfseries Ground truth}\par
Lorexa,\allowbreak{} z = -\allowbreak{}1.\allowbreak{}60.\allowbreak{}\par
\end{workedshared}

\begin{workedlaneheader}{workedFailFrame}{GHCP --- NO MEMORY}
config:\allowbreak{} GHCP (No Memory)\allowbreak{}\par
status:\allowbreak{} failure | criteria:\allowbreak{} 0/\allowbreak{}2 | primary reward:\allowbreak{} 0.\allowbreak{}00 | fractional reward:\allowbreak{} 0.\allowbreak{}00 | tool calls:\allowbreak{} 96\par
cost:\allowbreak{} \$4.\allowbreak{}1622 | in:\allowbreak{} 3.\allowbreak{}81M | out:\allowbreak{} 30.\allowbreak{}2K | cached:\allowbreak{} 3.\allowbreak{}46M\par
\end{workedlaneheader}
\begin{workedlanesection}{workedFailFrame}
MEMORY
\end{workedlanesection}
\begin{workedsystembubble}{workedResultBg}{workedFailFrame}{memory retrieval}
No memory is available to the stateless baseline.\allowbreak{}\par
\end{workedsystembubble}
\begin{workedlanesection}{workedFailFrame}
SELECTED AGENT CALLS
\end{workedlanesection}
\begin{workedagentbubble}{workedFailFrame}{agent \textperiodcentered{} archipelago-\allowbreak{}filesystem\_\allowbreak{}server\_\allowbreak{}list\_\allowbreak{}files \textperiodcentered{} representative discovery}
Recorded top-\allowbreak{}level call 1 of 31 | 0 before | 30 after\par
\smallskip
\{"path":\allowbreak{} "/\allowbreak{}"\}\allowbreak{}\par
\end{workedagentbubble}
\begin{workedsystembubble}{workedResultBg}{workedFailFrame}{env \textperiodcentered{} observation \textperiodcentered{} representative discovery}
'4.\allowbreak{} Received From Client' (folder)\allowbreak{}\par
'2.\allowbreak{} Analysis' (folder)\allowbreak{}\par
'1.\allowbreak{} Presentations' (folder)\allowbreak{}\par
'3.\allowbreak{} Research' (folder)\allowbreak{}\par
'0.\allowbreak{} Admin' (folder)\allowbreak{}\par
\smallskip
\end{workedsystembubble}
\begin{workedagentbubble}{workedFailFrame}{agent \textperiodcentered{} archipelago-\allowbreak{}filesystem\_\allowbreak{}server\_\allowbreak{}get\_\allowbreak{}directory\_\allowbreak{}tree \textperiodcentered{} representative discovery}
Recorded top-\allowbreak{}level call 2 of 31 | 1 before | 29 after\par
\smallskip
\{"path":\allowbreak{} "/\allowbreak{}",\allowbreak{} "max\_\allowbreak{}depth":\allowbreak{} 2,\allowbreak{} "include\_\allowbreak{}files":\allowbreak{} true,\allowbreak{} "show\_\allowbreak{}size":\allowbreak{} false\}\allowbreak{}\par
\end{workedagentbubble}
\begin{workedsystembubble}{workedResultBg}{workedFailFrame}{env \textperiodcentered{} observation \textperiodcentered{} representative discovery}
/\allowbreak{}\par
+-- 0.\allowbreak{} Admin/\allowbreak{}\par
|   +-- Meeting Notes/\allowbreak{}\par
+-- 1.\allowbreak{} Presentations/\allowbreak{}\par
|   +-- Decks/\allowbreak{}\par
|   +-- Outline/\allowbreak{}\par
+-- 2.\allowbreak{} Analysis/\allowbreak{}\par
|   +-- Archive/\allowbreak{}\par
|   +-- Site Ops/\allowbreak{}\par
|   +-- Supply Chain/\allowbreak{}\par
|   +-- Impact vs Comps High Level Metrics.\allowbreak{}xlsx\par
|   +-- Impact\_\allowbreak{}SG\&A Competitor Analysis.\allowbreak{}xlsx\par
|   +-- UPDATED\_\allowbreak{}Analysis -\allowbreak{} Impact\_\allowbreak{}US\_\allowbreak{}MarketAccessAgency\_\allowbreak{}Fees\_\allowbreak{}2020\_\allowbreak{}2024\_\allowbreak{}REVISED.\allowbreak{}xlsx\par
|   +-- UPDATED\_\allowbreak{}Analysis -\allowbreak{} Sales\_\allowbreak{}Force\_\allowbreak{}Comp\_\allowbreak{}2022\_\allowbreak{}2024\_\allowbreak{}CORRECTED.\allowbreak{}xlsx\par
|   +-- UPDATED\_\allowbreak{}Impact\_\allowbreak{}Campaign\_\allowbreak{}Analysis.\allowbreak{}xlsx\par
|   +-- UPDATED\_\allowbreak{}Impact\_\allowbreak{}RealEstate\_\allowbreak{}Analysi\par
\smallskip
[Observation capped at 500 source characters;\allowbreak{} original length:\allowbreak{} 630 source characters.\allowbreak{}]\allowbreak{}\par
\end{workedsystembubble}
\begin{workedagentbubble}{workedFailFrame}{agent \textperiodcentered{} archipelago-\allowbreak{}code\_\allowbreak{}execution\_\allowbreak{}server\_\allowbreak{}code\_\allowbreak{}exec \textperiodcentered{} final computation attempt (incorrect)\allowbreak{}}
Recorded top-\allowbreak{}level call 31 of 31 | 30 before | 0 after\par
\smallskip
\{"request":\allowbreak{} \{"code":\allowbreak{} "python -\allowbreak{} \textless{}\textless{}'PY'\textbackslash{}\allowbreak{}nimport pandas as pd\textbackslash{}\allowbreak{}nfrom openpyxl import load\_\allowbreak{}workbook\textbackslash{}\allowbreak{}n\# revenue\textbackslash{}\allowbreak{}nrev = 8324082.\allowbreak{}0\textbackslash{}\allowbreak{}n\# headcount data\textbackslash{}\allowbreak{}npath='/\allowbreak{}filesystem/\allowbreak{}4.\allowbreak{} Received From Client/\allowbreak{}Impact Therapeutics/\allowbreak{}Talent\_\allowbreak{}Comp\_\allowbreak{}Turnover\_\allowbreak{}AllSites (1)\allowbreak{}.\allowbreak{}xlsx'\textbackslash{}\allowbreak{}nwb=load\_\allowbreak{}workbook(path,\allowbreak{} read\_\allowbreak{}only=True,\allowbreak{} data\_\allowbreak{}only=True)\allowbreak{}\textbackslash{}\allowbreak{}nrows=list(wb['Overt\allowbreak{}ime\_\allowbreak{}and\_\allowbreak{}Vacancy\_\allowbreak{}Log']\allowbreak{}.\allowbreak{}values)\allowbreak{}\textbackslash{}\allowbreak{}nwb.\allowbreak{}close()\allowbreak{}\textbackslash{}\allowbreak{}ndf=pd.\allowbreak{}DataFrame(rows[1:\allowbreak{}]\allowbreak{},\allowbreak{} columns=rows[0]\allowbreak{})\allowbreak{}[['month',\allowbreak{}'site\_\allowbreak{}id',\allowbreak{}'headcount\_\allowbreak{}actual']\allowbreak{}]\allowbreak{}.\allowbreak{}dropna()\allowbreak{}\textbackslash{}\allowbreak{}nsite\_\allowbreak{}count=df['site\_\allowbreak{}id']\allowbreak{}.\allowbreak{}nunique()\allowbreak{}\textbackslash{}\allowbreak{}nmonthly\_\allowbreak{}revenue\_\allowbreak{}per\_\allowbreak{}site=rev/\allowbreak{}site\_\allowbreak{}count/\allowbreak{}12\textbackslash{}\allowbreak{}nsite\_\allowbreak{}metrics=df.\allowbreak{}assign(monthly\_\allowbreak{}rev\_\allowbreak{}per\_\allowbreak{}head=monthly\_\allowbreak{}revenue\_\allowbreak{}per\_\allowbreak{}\par
\ldots{}[truncated 414 chars]\allowbreak{}\par
\end{workedagentbubble}
\begin{workedsystembubble}{workedResultBg}{workedFailFrame}{env \textperiodcentered{} observation \textperiodcentered{} final computation attempt (incorrect)\allowbreak{}}
\{\par
  "success":\allowbreak{} true,\allowbreak{}\par
  "output":\allowbreak{} "site\_\allowbreak{}count 5\textbackslash{}\allowbreak{}nmonthly\_\allowbreak{}revenue\_\allowbreak{}per\_\allowbreak{}site 138734.\allowbreak{}69999999998\textbackslash{}\allowbreak{}n\{'Darcylis':\allowbreak{} 289.\allowbreak{}38303158338863,\allowbreak{} 'Lorexa':\allowbreak{} 266.\allowbreak{}9690189637408,\allowbreak{} 'Noralix':\allowbreak{} 177.\allowbreak{}86499999999998,\allowbreak{} 'Papinex-\allowbreak{}9':\allowbreak{} 182.\allowbreak{}66596483349187,\allowbreak{} 'Strevalent-\allowbreak{}20':\allowbreak{} 200.\allowbreak{}63943080874193\}\allowbreak{}\textbackslash{}\allowbreak{}nmean 223.\allowbreak{}50448923787266\textbackslash{}\allowbreak{}nsd\_\allowbreak{}sample 51.\allowbreak{}24135692058095\textbackslash{}\allowbreak{}nz \{'Darcylis':\allowbreak{} 1.\allowbreak{}2856517919230985,\allowbreak{} 'Noralix':\allowbreak{} -\allowbreak{}0.\allowbreak{}8906768278718573,\allowbreak{} 'Lorexa':\allowbreak{} 0.\allowbreak{}8482314352688562,\allowbreak{} 'Papinex-\allowbreak{}9':\allowbreak{} -\allowbreak{}0.\allowbreak{}7969836643412171,\allowbreak{} 'Strevalent-\allowbreak{}20':\allowbreak{} -\allowbreak{}0.\allowbreak{}44622273497888265\}\allowbreak{}\textbackslash{}\allowbreak{}n"\par
\}\allowbreak{}\par
\end{workedsystembubble}
\begin{workedsystembubble}{workedResultBg}{workedFailFrame}{failed-\allowbreak{}path note}
No valid decisive cross-\allowbreak{}company computation emerged;\allowbreak{} the trajectory ended with the incorrect Darcylis,\allowbreak{} z=1.\allowbreak{}29 result.\allowbreak{}\par
\end{workedsystembubble}
\begin{workedlanesection}{workedFailFrame}
OTHER RECORDED CALLS BY BUCKET
\end{workedlanesection}
\begin{workedsystembubble}{workedResultBg}{workedFailFrame}{other recorded calls}
Other recorded calls (not expanded)\allowbreak{}:\allowbreak{} filesystem x11;\allowbreak{} spreadsheet x7;\allowbreak{} code execution x8;\allowbreak{} mail x2.\allowbreak{} Accounting:\allowbreak{} 31 recorded;\allowbreak{} 3 shown;\allowbreak{} 28 not expanded.\allowbreak{} 96 total tool calls.\allowbreak{}\par
\end{workedsystembubble}
\begin{workedlanesection}{workedFailFrame}
FINAL ANSWER AND GRADING
\end{workedlanesection}
\begin{workedagentbubble}{workedFailFrame}{submitted final answer}
\textless{}TEXT\_\allowbreak{}RESPONSE\textgreater{}\par
Using the available site-\allowbreak{}level monthly US ops data in the workspace,\allowbreak{} the Impact site with the highest absolute z score is:\allowbreak{}\par
\smallskip
**SiteID:\allowbreak{}** Darcylis  \par
**z score:\allowbreak{}** **1.\allowbreak{}29**\par
\smallskip
I could not locate a combined cross-\allowbreak{}company site-\allowbreak{}level operational attachment,\allowbreak{} so this result is based on the accessible Impact site-\allowbreak{}level monthly headcount file and Impact's 2024 US revenue,\allowbreak{} using **sample** standard deviation.\allowbreak{} Sources:\allowbreak{} [Impact vs Comps High Level Metrics.\allowbreak{}xlsx (`Impact US`)\allowbreak{}]\allowbreak{}(/\allowbreak{}2.\allowbreak{} Analysis/\allowbreak{}Impact vs Comps High Level Metrics.\allowbreak{}xlsx)\allowbreak{},\allowbreak{} [Talent\_\allowbreak{}Comp\_\allowbreak{}Turnover\_\allowbreak{}AllSites (1)\allowbreak{}.\allowbreak{}xlsx (`Overtime\_\allowbreak{}and\_\allowbreak{}Vacancy\_\allowbreak{}Log`)\allowbreak{}]\allowbreak{}(/\allowbreak{}4.\allowbreak{} Received From Client/\allowbreak{}Impact Therapeutics/\allowbreak{}Talent\_\allowbreak{}Comp\_\allowbreak{}Turnover\_\allowbreak{}AllSites (1)\allowbreak{}.\allowbreak{}xlsx)\allowbreak{} \textless{}DONE\textgreater{}\par
\textless{}/\allowbreak{}TEXT\_\allowbreak{}RESPONSE\textgreater{}\par
\smallskip
\textless{}FILE\_\allowbreak{}CHANGES\textgreater{}\par
\textless{}ARTIFACT id="1" type="sheet" change="deleted"\textgreater{}\par
  \textless{}path\textgreater{}Manufacturing\_\allowbreak{}Site\_\allowbreak{}Operations\_\allowbreak{}2020\_\allowbreak{}2024.\allowbreak{}xlsx\textless{}/\allowbreak{}path\textgreater{}\par
  \textless{}title\textgreater{}Manufacturing\allowbreak{}\_\allowbreak{}Site\_\allowbreak{}Operations\_\allowbreak{}2\textless{}/\allowbreak{}title\textgreater{}\par
  \textless{}sub\_\allowbreak{}index\textgreater{}1\textless{}/\allowbreak{}sub\_\allowbreak{}index\textgreater{}\par
  \textless{}deleted\_\allowbreak{}content\textgreater{}\par
RecordID,\allowbreak{}Company,\allowbreak{}SiteID,\allowbreak{}Year,\allowbreak{}Month,\allowbreak{}ProductType,\allowbreak{}BatchesProduced,\allowbreak{}BatchesPassed,\allowbreak{}EquipmentType,\allowbreak{}EquipmentUtilization\allowbreak{},\allowbreak{}HeadcountFTE,\allowbreak{}CapexInvested,\allowbreak{}MaintenanceHours,\allowbreak{}EnergyConsumptionMWh\allowbreak{},\allowbreak{}WasteKg\par
1,\allowbreak{}ImpactTherapeutics,\allowbreak{}Darcylis,\allowbreak{}2020,\allowbreak{}1,\allowbreak{}Gene Therapy,\allowbreak{}9,\allowbreak{}8,\allowbreak{}Fill-\allowbreak{}Finish,\allowbreak{}0.\allowbreak{}862,\allowbreak{}31.\allowbreak{}7,\allowbreak{}250625,\allowbreak{}56,\allowbreak{}421.\allowbreak{}5,\allowbreak{}800.\allowbreak{}7\par
2,\allowbreak{}ImpactTherapeutics,\allowbreak{}Darcylis,\allowbreak{}2020,\allowbreak{}2,\allowbreak{}Gene Therapy,\allowbreak{}10,\allowbreak{}8,\allowbreak{}Fermentation,\allowbreak{}0.\allowbreak{}938,\allowbreak{}39.\allowbreak{}5,\allowbreak{}145553,\allowbreak{}69.\allowbreak{}1,\allowbreak{}228.\allowbreak{}4,\allowbreak{}747.\allowbreak{}6\par
3,\allowbreak{}ImpactTherapeutics,\allowbreak{}Darcylis,\allowbreak{}2020,\allowbreak{}3,\allowbreak{}Biologic,\allowbreak{}8,\allowbreak{}6,\allowbreak{}Fermentation,\allowbreak{}0.\allowbreak{}76,\allowbreak{}25.\allowbreak{}2,\allowbreak{}71000,\allowbreak{}195.\allowbreak{}8,\allowbreak{}262.\allowbreak{}9,\allowbreak{}363.\allowbreak{}1\par
4,\allowbreak{}ImpactTherapeutics,\allowbreak{}Darcylis,\allowbreak{}2020,\allowbreak{}4,\allowbreak{}Gene Therapy,\allowbreak{}5,\allowbreak{}4,\allowbreak{}Quality Control,\allowbreak{}0.\allowbreak{}737,\allowbreak{}40.\allowbreak{}4,\allowbreak{}356138,\allowbreak{}112.\allowbreak{}1,\allowbreak{}109.\allowbreak{}3,\allowbreak{}1896\par
5,\allowbreak{}ImpactTherapeutics,\allowbreak{}Darcylis,\allowbreak{}2020,\allowbreak{}5,\allowbreak{}Small Molecule,\allowbreak{}4,\allowbreak{}3,\allowbreak{}Fermentation,\allowbreak{}0.\allowbreak{}642,\allowbreak{}20,\allowbreak{}357469,\allowbreak{}137.\allowbreak{}6,\allowbreak{}683.\allowbreak{}2,\allowbreak{}512.\allowbreak{}1\par
6,\allowbreak{}ImpactTherapeutics,\allowbreak{}Darcylis,\allowbreak{}2020,\allowbreak{}6,\allowbreak{}Gene Therapy,\allowbreak{}3,\allowbreak{}2,\allowbreak{}Quality Control,\allowbreak{}0.\allowbreak{}675,\allowbreak{}29.\allowbreak{}2,\allowbreak{}296020,\allowbreak{}69.\allowbreak{}6,\allowbreak{}778.\allowbreak{}7,\allowbreak{}1595.\allowbreak{}2\par
7,\allowbreak{}ImpactTherapeutics,\allowbreak{}Darcylis,\allowbreak{}2020,\allowbreak{}7,\allowbreak{}Small Molecule,\allowbreak{}4,\allowbreak{}3,\allowbreak{}Purification,\allowbreak{}0.\allowbreak{}841,\allowbreak{}22.\allowbreak{}8,\allowbreak{}306700,\allowbreak{}123.\allowbreak{}3,\allowbreak{}772.\allowbreak{}8,\allowbreak{}1720.\allowbreak{}2\par
8,\allowbreak{}ImpactTherapeutics,\allowbreak{}Darcylis,\allowbreak{}2020,\allowbreak{}8,\allowbreak{}Vaccine,\allowbreak{}4,\allowbreak{}3,\allowbreak{}Purification,\allowbreak{}0.\allowbreak{}936,\allowbreak{}32,\allowbreak{}174200,\allowbreak{}87.\allowbreak{}4,\allowbreak{}215.\allowbreak{}7,\allowbreak{}228.\allowbreak{}1\par
9,\allowbreak{}ImpactTherapeutics,\allowbreak{}Darcylis,\allowbreak{}2020,\allowbreak{}9,\allowbreak{}Gene Therapy,\allowbreak{}11,\allowbreak{}9,\allowbreak{}Quality Control,\allowbreak{}0.\allowbreak{}552,\allowbreak{}38.\allowbreak{}9,\allowbreak{}368086,\allowbreak{}156.\allowbreak{}6,\allowbreak{}639.\allowbreak{}9,\allowbreak{}333.\allowbreak{}3\par
10,\allowbreak{}ImpactTherapeutics,\allowbreak{}Darcylis,\allowbreak{}2020,\allowbreak{}10,\allowbreak{}Small Molecule,\allowbreak{}9,\allowbreak{}6,\allowbreak{}Fill-\allowbreak{}Finish,\allowbreak{}0.\allowbreak{}89,\allowbreak{}26.\allowbreak{}8,\allowbreak{}92935,\allowbreak{}99.\allowbreak{}3,\allowbreak{}568.\allowbreak{}2,\allowbreak{}1398.\allowbreak{}7\par
11,\allowbreak{}ImpactTherapeutics,\allowbreak{}Darcylis,\allowbreak{}2020,\allowbreak{}11,\allowbreak{}Small Molecule,\allowbreak{}10,\allowbreak{}9,\allowbreak{}Fermentation,\allowbreak{}0.\allowbreak{}703,\allowbreak{}44.\allowbreak{}1,\allowbreak{}432011,\allowbreak{}155.\allowbreak{}5,\allowbreak{}265.\allowbreak{}2,\allowbreak{}660.\allowbreak{}9\par
12,\allowbreak{}ImpactTherapeutics,\allowbreak{}Darcylis,\allowbreak{}2020,\allowbreak{}12,\allowbreak{}Vaccine,\allowbreak{}11,\allowbreak{}9,\allowbreak{}Purification,\allowbreak{}0.\allowbreak{}726,\allowbreak{}18.\allowbreak{}7,\allowbreak{}453094,\allowbreak{}116.\allowbreak{}1,\allowbreak{}494.\allowbreak{}3,\allowbreak{}1451.\allowbreak{}9\par
13,\allowbreak{}ImpactTherapeutics,\allowbreak{}Darcylis,\allowbreak{}2021,\allowbreak{}1,\allowbreak{}Vaccine,\allowbreak{}7,\allowbreak{}6,\allowbreak{}Fill-\allowbreak{}Finish,\allowbreak{}0.\allowbreak{}852,\allowbreak{}19.\allowbreak{}6,\allowbreak{}84641,\allowbreak{}86.\allowbreak{}4,\allowbreak{}212.\allowbreak{}9,\allowbreak{}1873.\allowbreak{}5\par
14,\allowbreak{}ImpactTherapeutics,\allowbreak{}Darcylis,\allowbreak{}2021,\allowbreak{}2,\allowbreak{}Biologic,\allowbreak{}11,\allowbreak{}10,\allowbreak{}Quality Control,\allowbreak{}0.\allowbreak{}637,\allowbreak{}25.\allowbreak{}7,\allowbreak{}447476,\allowbreak{}91.\allowbreak{}9,\allowbreak{}185.\allowbreak{}5,\allowbreak{}841.\allowbreak{}3\par
15,\allowbreak{}ImpactTherapeutics,\allowbreak{}Darcylis,\allowbreak{}2021,\allowbreak{}3,\allowbreak{}Biologic,\allowbreak{}3,\allowbreak{}2,\allowbreak{}Purification,\allowbreak{}0.\allowbreak{}55,\allowbreak{}23.\allowbreak{}6,\allowbreak{}187152,\allowbreak{}66.\allowbreak{}3,\allowbreak{}473.\allowbreak{}9,\allowbreak{}1072.\allowbreak{}7\par
16,\allowbreak{}ImpactTherapeutics,\allowbreak{}Darcylis,\allowbreak{}2021,\allowbreak{}4,\allowbreak{}Vaccine,\allowbreak{}3,\allowbreak{}2,\allowbreak{}Quality Control,\allowbreak{}0.\allowbreak{}927,\allowbreak{}22.\allowbreak{}7,\allowbreak{}283456,\allowbreak{}152.\allowbreak{}5,\allowbreak{}354.\allowbreak{}5,\allowbreak{}1949.\allowbreak{}2\par
17,\allowbreak{}ImpactTherapeutics,\allowbreak{}Darcylis,\allowbreak{}2021,\allowbreak{}5,\allowbreak{}Gene Therapy,\allowbreak{}6,\allowbreak{}4,\allowbreak{}Purification,\allowbreak{}0.\allowbreak{}835,\allowbreak{}16.\allowbreak{}9,\allowbreak{}498983,\allowbreak{}82.\allowbreak{}7,\allowbreak{}783.\allowbreak{}6,\allowbreak{}939.\allowbreak{}9\par
18,\allowbreak{}ImpactTherapeutics,\allowbreak{}Darcylis,\allowbreak{}2021,\allowbreak{}6,\allowbreak{}Biologic,\allowbreak{}9,\allowbreak{}7,\allowbreak{}Fermentation,\allowbreak{}0.\allowbreak{}646,\allowbreak{}16.\allowbreak{}8,\allowbreak{}270254,\allowbreak{}197.\allowbreak{}7,\allowbreak{}269.\allowbreak{}4,\allowbreak{}1409.\allowbreak{}8\par
19,\allowbreak{}ImpactTherapeutics,\allowbreak{}Darcylis,\allowbreak{}2021,\allowbreak{}7,\allowbreak{}V\par
\ldots{}[truncated 160239 chars]\allowbreak{}\par
\end{workedagentbubble}
\begin{workedsystembubble}{workedResultBg}{workedFailFrame}{grader criteria}
0 of 2 passed\par
\end{workedsystembubble}
\begin{workedsystembubble}{workedFailBg}{workedFailFrame}{grader criterion \textperiodcentered{} FAIL}
Name:\allowbreak{} States the Impact site with the highest absolute z score is Lorexa\par
\smallskip
Evidence:\allowbreak{} In the agent's `TEXT\_\allowbreak{}RESPONSE`,\allowbreak{} it states:\allowbreak{} "**SiteID:\allowbreak{}** Darcylis" and "**z score:\allowbreak{}** **1.\allowbreak{}29**".\allowbreak{} The only file artifact provided,\allowbreak{} ARTIFACT 1,\allowbreak{} is a deleted sheet containing ImpactTherapeutics data for sites such as Darcylis,\allowbreak{} Lorexa,\allowbreak{} Strevalent,\allowbreak{} and Noralix,\allowbreak{} but no artifact states that Lorexa is the highest-\allowbreak{}absolute-\allowbreak{}z-\allowbreak{}score Impact site.\allowbreak{}\par
\smallskip
Assessment:\allowbreak{} The criterion specifically asks to "States the Impact site with the highest absolute z score is Lorexa.\allowbreak{}" This is not met because the agent explicitly named Darcylis instead of Lorexa in its response.\allowbreak{} Therefore,\allowbreak{} the criterion fails.\allowbreak{}\par
\end{workedsystembubble}
\begin{workedsystembubble}{workedFailBg}{workedFailFrame}{grader criterion \textperiodcentered{} FAIL}
Name:\allowbreak{} States that the z score of Lorexa is -\allowbreak{}1.\allowbreak{}60\par
\smallskip
Evidence:\allowbreak{} In the agent's final text response,\allowbreak{} it states:\allowbreak{} "**SiteID:\allowbreak{}** Darcylis" and "**z score:\allowbreak{}** **1.\allowbreak{}29**,\allowbreak{}" which does not mention Lorexa or -\allowbreak{}1.\allowbreak{}60.\allowbreak{} The only artifact provided,\allowbreak{} `Manufacturing\_\allowbreak{}Site\_\allowbreak{}Operations\_\allowbreak{}2020\_\allowbreak{}2024.\allowbreak{}xlsx` (ARTIFACT 1)\allowbreak{},\allowbreak{} contains raw site rows including `ImpactTherapeutics,\allowbreak{}Lorexa` in the deleted content,\allowbreak{} but no z-\allowbreak{}score calculation or statement for Lorexa.\allowbreak{}\par
\smallskip
Assessment:\allowbreak{} The criterion specifically asks to "state that the z score of Lorexa is -\allowbreak{}1.\allowbreak{}60.\allowbreak{}" This was not met:\allowbreak{} the response gives a different site (Darcylis)\allowbreak{} and a different z score (1.\allowbreak{}29)\allowbreak{},\allowbreak{} and the artifact does not provide a Lorexa z score of -\allowbreak{}1.\allowbreak{}60.\allowbreak{} Therefore,\allowbreak{} the criterion is false.\allowbreak{}\par
\end{workedsystembubble}
\par\vspace{6pt}
\begin{workedlaneheader}{workedMemoryFrame}{GHCP + Mem}
config:\allowbreak{} GHCP + Mem\par
status:\allowbreak{} success | criteria:\allowbreak{} 2/\allowbreak{}2 | primary reward:\allowbreak{} 0.\allowbreak{}89 | fractional reward:\allowbreak{} 0.\allowbreak{}89 | tool calls:\allowbreak{} 11\par
cost:\allowbreak{} \$0.\allowbreak{}3756 | in:\allowbreak{} 294.\allowbreak{}3K | out:\allowbreak{} 4.\allowbreak{}9K | cached:\allowbreak{} 267.\allowbreak{}9K\par
\end{workedlaneheader}
\begin{workedlanesection}{workedMemoryFrame}
MEMORY
\end{workedlanesection}
\begin{workedagentbubble}{workedMemoryFrame}{agent \textperiodcentered{} memory\_\allowbreak{}read}
\{"query":\allowbreak{} "Calculate z score of US 2024 Average Monthly Revenue per Head for Impact sites from attached operational data file,\allowbreak{} using yearly revenue allocated equally across sites and months;\allowbreak{} need highest absolute z-\allowbreak{}score Impact site and SiteID.\allowbreak{}",\allowbreak{} "k":\allowbreak{} 10\}\allowbreak{}\par
\end{workedagentbubble}
\begin{workedsystembubble}{workedResultBg}{workedMemoryFrame}{memory \textperiodcentered{} complete logical payload}
memory\_\allowbreak{}read:\allowbreak{} showing 11/\allowbreak{}21 matched entries (query asked for top-\allowbreak{}25;\allowbreak{} index holds 21 total)\allowbreak{}.\allowbreak{}\par
\smallskip
\#1 [schema]\allowbreak{}\par
applies\_\allowbreak{}to:\allowbreak{} benchmark spreadsheet tasks combining Manufacturing\_\allowbreak{}Site\_\allowbreak{}Operations\_\allowbreak{}2020\_\allowbreak{}2024.\allowbreak{}xlsx with company PnL US sheets for 2024 site revenue-\allowbreak{}per-\allowbreak{}head comparisons\par
lemma:\allowbreak{} When a prompt allocates yearly US revenue equally across 2024 sites and months,\allowbreak{} compute company monthly site revenue as annual US Total Revenue $\div$ 2024 site count $\div$ 12,\allowbreak{} divide each site-\allowbreak{}month by HeadcountFTE to get revenue per head,\allowbreak{} then average those monthly ratios by site before comparing sites.\allowbreak{}\par
\smallskip
\#2 [rule]\allowbreak{}\par
applies\_\allowbreak{}to:\allowbreak{} Manufacturing\_\allowbreak{}Site\_\allowbreak{}Operations\_\allowbreak{}2020\_\allowbreak{}2024.\allowbreak{}xlsx + PnL what-\allowbreak{}if analyses that ask Impact's lowest-\allowbreak{}revenue-\allowbreak{}per-\allowbreak{}head site to use the highest-\allowbreak{}revenue site's utilization across Impact and six peers\par
lemma:\allowbreak{} Treat the benchmark as the highest-\allowbreak{}revenue-\allowbreak{}per-\allowbreak{}head site across Impact plus all six peers (not Impact-\allowbreak{}only)\allowbreak{},\allowbreak{} use that site's average 2024 EquipmentUtilization\allowbreak{} rounded to two decimals,\allowbreak{} set each target month's utilization to max(current,\allowbreak{} benchmark)\allowbreak{},\allowbreak{} and scale revenue by new\_\allowbreak{}utilization $\div$ current\_\allowbreak{}utilization.\allowbreak{}\par
\smallskip
\#3 [rule]\allowbreak{}\par
applies\_\allowbreak{}to:\allowbreak{} Rebasing 2024 Impact cost-\allowbreak{}sa\par
\ldots{}[truncated 3800 chars]\allowbreak{}\par
\end{workedsystembubble}
\begin{workedlanesection}{workedMemoryFrame}
SELECTED AGENT CALLS
\end{workedlanesection}
\begin{workedagentbubble}{workedMemoryFrame}{agent \textperiodcentered{} archipelago-\allowbreak{}filesystem\_\allowbreak{}server\_\allowbreak{}list\_\allowbreak{}files \textperiodcentered{} representative discovery}
Recorded top-\allowbreak{}level call 1 of 11 | 0 before | 10 after\par
\smallskip
\{"path":\allowbreak{} "/\allowbreak{}"\}\allowbreak{}\par
\end{workedagentbubble}
\begin{workedsystembubble}{workedResultBg}{workedMemoryFrame}{env \textperiodcentered{} observation \textperiodcentered{} representative discovery}
'4.\allowbreak{} Received From Client' (folder)\allowbreak{}\par
'2.\allowbreak{} Analysis' (folder)\allowbreak{}\par
'Manufacturing\_\allowbreak{}Site\_\allowbreak{}Operations\_\allowbreak{}2020\_\allowbreak{}2024.\allowbreak{}xlsx' (application/\allowbreak{}vnd.\allowbreak{}openxmlformats-\allowbreak{}officedocument.\allowbreak{}spreadsheetml.\allowbreak{}sheet file)\allowbreak{} 156159 bytes\par
'1.\allowbreak{} Presentations' (folder)\allowbreak{}\par
'3.\allowbreak{} Research' (folder)\allowbreak{}\par
'0.\allowbreak{} Admin' (folder)\allowbreak{}\par
\smallskip
\end{workedsystembubble}
\begin{workedagentbubble}{workedMemoryFrame}{agent \textperiodcentered{} archipelago-\allowbreak{}sheets\_\allowbreak{}server\_\allowbreak{}sheets \textperiodcentered{} representative discovery}
Recorded top-\allowbreak{}level call 2 of 11 | 1 before | 9 after\par
\smallskip
\{"request":\allowbreak{} \{"action":\allowbreak{} "list\_\allowbreak{}tabs",\allowbreak{} "file\_\allowbreak{}path":\allowbreak{} "/\allowbreak{}Manufacturing\_\allowbreak{}Site\_\allowbreak{}Operations\_\allowbreak{}2020\_\allowbreak{}2024.\allowbreak{}xlsx"\}\allowbreak{}\}\allowbreak{}\par
\end{workedagentbubble}
\begin{workedsystembubble}{workedResultBg}{workedMemoryFrame}{env \textperiodcentered{} observation \textperiodcentered{} representative discovery}
\{\par
  "action":\allowbreak{} "list\_\allowbreak{}tabs",\allowbreak{}\par
  "error":\allowbreak{} null,\allowbreak{}\par
  "help":\allowbreak{} null,\allowbreak{}\par
  "read\_\allowbreak{}tab":\allowbreak{} null,\allowbreak{}\par
  "read\_\allowbreak{}csv":\allowbreak{} null,\allowbreak{}\par
  "create":\allowbreak{} null,\allowbreak{}\par
  "delete":\allowbreak{} null,\allowbreak{}\par
  "list\_\allowbreak{}tabs":\allowbreak{} \{\par
    "raw\_\allowbreak{}output":\allowbreak{} "\{'worksheets':\allowbreak{} [\{'name':\allowbreak{} 'Manufacturing\_\allowbreak{}Site\_\allowbreak{}Operations\_\allowbreak{}2',\allowbreak{} 'index':\allowbreak{} 0,\allowbreak{} 'row\_\allowbreak{}count':\allowbreak{} 1681,\allowbreak{} 'column\_\allowbreak{}count':\allowbreak{} 15\}\allowbreak{}]\allowbreak{}\}\allowbreak{}"\par
  \}\allowbreak{},\allowbreak{}\par
  "add\_\allowbreak{}tab":\allowbreak{} null,\allowbreak{}\par
  "delete\_\allowbreak{}tab":\allowbreak{} null,\allowbreak{}\par
  "edit":\allowbreak{} null,\allowbreak{}\par
  "add\_\allowbreak{}content":\allowbreak{} null,\allowbreak{}\par
  "delete\_\allowbreak{}content":\allowbreak{} null,\allowbreak{}\par
  "create\_\allowbreak{}chart":\allowbreak{} null,\allowbreak{}\par
  "filter\_\allowbreak{}tab":\allowbreak{} null\par
\}\allowbreak{}\par
\end{workedsystembubble}
\begin{workedagentbubble}{workedMemoryFrame}{agent \textperiodcentered{} archipelago-\allowbreak{}code\_\allowbreak{}execution\_\allowbreak{}server\_\allowbreak{}code\_\allowbreak{}exec \textperiodcentered{} decisive computation}
Recorded top-\allowbreak{}level call 11 of 11 | 10 before | 0 after\par
\smallskip
\{"request":\allowbreak{} \{"code":\allowbreak{} "python -\allowbreak{} \textless{}\textless{}'PY'\textbackslash{}\allowbreak{}nimport pandas as pd\textbackslash{}\allowbreak{}nimport openpyxl as ox\textbackslash{}\allowbreak{}nops = pd.\allowbreak{}read\_\allowbreak{}excel('/\allowbreak{}filesystem/\allowbreak{}Manufacturing\_\allowbreak{}Site\_\allowbreak{}Operations\_\allowbreak{}2020\_\allowbreak{}2024.\allowbreak{}xlsx')\allowbreak{}\textbackslash{}\allowbreak{}nops = ops[ops.\allowbreak{}Year==2024]\allowbreak{}.\allowbreak{}copy()\allowbreak{}\textbackslash{}\allowbreak{}nrev = \{\textbackslash{}\allowbreak{}n'ImpactTherapeutics\allowbreak{}':\allowbreak{}8324082.\allowbreak{}0,\allowbreak{}'BayGenea':\allowbreak{}27654678.\allowbreak{}0,\allowbreak{}'Fyrix':\allowbreak{}32490052.\allowbreak{}0,\allowbreak{}'HelioZenon':\allowbreak{}26234208.\allowbreak{}0,\allowbreak{}'Abvuera':\allowbreak{}30440958.\allowbreak{}0,\allowbreak{}'Meridon':\allowbreak{}31740145.\allowbreak{}0,\allowbreak{}'Renueva':\allowbreak{}7584932.\allowbreak{}0\}\allowbreak{}\textbackslash{}\allowbreak{}nsite\_\allowbreak{}counts = ops.\allowbreak{}groupby('Company')\allowbreak{}['SiteID']\allowbreak{}.\allowbreak{}nunique()\allowbreak{}.\allowbreak{}to\_\allowbreak{}dict()\allowbreak{}\textbackslash{}\allowbreak{}nops['annual\_\allowbreak{}site\_\allowbreak{}rev']\allowbreak{} = ops['Company']\allowbreak{}.\allowbreak{}map(\{k:\allowbreak{}v/\allowbreak{}site\_\allowbreak{}counts[k]\allowbreak{} for k,\allowbreak{}v in rev.\allowbreak{}items()\allowbreak{}\}\allowbreak{})\allowbreak{}\textbackslash{}\allowbreak{}n\# Method A average monthly ratio\textbackslash{}\allowbreak{}nops['m\_\allowbreak{}site\_\allowbreak{}rev']\allowbreak{} = ops['annual\_\allowbreak{}site\_\allowbreak{}rev'\par
\ldots{}[truncated 989 chars]\allowbreak{}\par
\end{workedagentbubble}
\begin{workedsystembubble}{workedResultBg}{workedMemoryFrame}{env \textperiodcentered{} observation \textperiodcentered{} decisive computation}
\{"success":\allowbreak{}true,\allowbreak{}"output":\allowbreak{}"\textbackslash{}\allowbreak{}n A\textbackslash{}\allowbreak{}n           Company     SiteID     metricA         z      abs\textbackslash{}\allowbreak{}nImpactTherapeutics     Lorexa 5000.\allowbreak{}669739 -\allowbreak{}1.\allowbreak{}598424 1.\allowbreak{}598424\textbackslash{}\allowbreak{}nImpactTherapeutics    Noralix 5475.\allowbreak{}895103 -\allowbreak{}1.\allowbreak{}544280 1.\allowbreak{}544280\textbackslash{}\allowbreak{}nImpactTherapeutics    Papinex 5823.\allowbreak{}362205 -\allowbreak{}1.\allowbreak{}504692 1.\allowbreak{}504692\textbackslash{}\allowbreak{}nImpactTherapeutics Strevalent 5954.\allowbreak{}572667 -\allowbreak{}1.\allowbreak{}489743 1.\allowbreak{}489743\textbackslash{}\allowbreak{}nImpactTherapeutics   Darcylis 6198.\allowbreak{}583404 -\allowbreak{}1.\allowbreak{}461942 1.\allowbreak{}461942\textbackslash{}\allowbreak{}n\textbackslash{}\allowbreak{}n B\textbackslash{}\allowbreak{}n           Company     SiteID     avg\_\allowbreak{}h  annual\_\allowbreak{}site\_\allowbreak{}rev     metricB         z      abs\textbackslash{}\allowbreak{}nI\par
\smallskip
[Observation capped at 500 source characters;\allowbreak{} original length:\allowbreak{} 1167 source characters.\allowbreak{}]\allowbreak{}\par
\end{workedsystembubble}
\begin{workedlanesection}{workedMemoryFrame}
OTHER RECORDED CALLS BY BUCKET
\end{workedlanesection}
\begin{workedsystembubble}{workedResultBg}{workedMemoryFrame}{other recorded calls}
Other recorded calls (not expanded)\allowbreak{}:\allowbreak{} filesystem x1;\allowbreak{} spreadsheet x3;\allowbreak{} code execution x4.\allowbreak{} Accounting:\allowbreak{} 11 recorded;\allowbreak{} 3 shown;\allowbreak{} 8 not expanded.\allowbreak{}\par
\end{workedsystembubble}
\begin{workedlanesection}{workedMemoryFrame}
FINAL ANSWER AND GRADING
\end{workedlanesection}
\begin{workedagentbubble}{workedMemoryFrame}{submitted final answer}
\textless{}TEXT\_\allowbreak{}RESPONSE\textgreater{}\par
SiteID:\allowbreak{} **Lorexa**  \par
z-\allowbreak{}score:\allowbreak{} **-\allowbreak{}1.\allowbreak{}60**\par
\smallskip
Calculated using `/\allowbreak{}Manufacturing\_\allowbreak{}Site\_\allowbreak{}Operations\_\allowbreak{}2020\_\allowbreak{}2024.\allowbreak{}xlsx` and the 2024 US `Total Revenue` from the following P\&L workbooks:\allowbreak{} [`/\allowbreak{}4.\allowbreak{} Received From Client/\allowbreak{}Impact Therapeutics/\allowbreak{}ImpactTherapeutics\_\allowbreak{}PnL\_\allowbreak{}Vertical\_\allowbreak{}2020\_\allowbreak{}2024.\allowbreak{}xlsx`]\allowbreak{}(\#)\allowbreak{},\allowbreak{} [`/\allowbreak{}4.\allowbreak{} Received From Client/\allowbreak{}Competition/\allowbreak{}Abvuera\_\allowbreak{}PnL\_\allowbreak{}Vertical\_\allowbreak{}2020\_\allowbreak{}2024\_\allowbreak{}Formatted.\allowbreak{}xlsx`]\allowbreak{}(\#)\allowbreak{},\allowbreak{} [`/\allowbreak{}4.\allowbreak{} Received From Client/\allowbreak{}Competition/\allowbreak{}BayGenea\_\allowbreak{}PnL\_\allowbreak{}Vertical\_\allowbreak{}2020\_\allowbreak{}2024.\allowbreak{}xlsx`]\allowbreak{}(\#)\allowbreak{},\allowbreak{} [`/\allowbreak{}4.\allowbreak{} Received From Client/\allowbreak{}Competition/\allowbreak{}Fyrix\_\allowbreak{}PnL\_\allowbreak{}Vertical\_\allowbreak{}2020\_\allowbreak{}2024\_\allowbreak{}Formatted.\allowbreak{}xlsx`]\allowbreak{}(\#)\allowbreak{},\allowbreak{} [`/\allowbreak{}4.\allowbreak{} Received From Client/\allowbreak{}Competition/\allowbreak{}HelioZenon\_\allowbreak{}PnL\_\allowbreak{}Vertical\_\allowbreak{}2020\_\allowbreak{}2024.\allowbreak{}xlsx`]\allowbreak{}(\#)\allowbreak{},\allowbreak{} [`/\allowbreak{}4.\allowbreak{} Received From Client/\allowbreak{}Competition/\allowbreak{}Meridon\_\allowbreak{}PnL\_\allowbreak{}Vertical\_\allowbreak{}2020\_\allowbreak{}2024.\allowbreak{}xlsx`]\allowbreak{}(\#)\allowbreak{},\allowbreak{} [`/\allowbreak{}4.\allowbreak{} Received From Client/\allowbreak{}Competition/\allowbreak{}Renueva\_\allowbreak{}PnL\_\allowbreak{}Vertical\_\allowbreak{}2020\_\allowbreak{}2024\_\allowbreak{}Formatted.\allowbreak{}xlsx`]\allowbreak{}(\#)\allowbreak{}.\allowbreak{} \textless{}DONE\textgreater{}\par
\textless{}/\allowbreak{}TEXT\_\allowbreak{}RESPONSE\textgreater{}\par
\end{workedagentbubble}
\begin{workedsystembubble}{workedResultBg}{workedMemoryFrame}{grader criteria}
2 of 2 passed\par
\end{workedsystembubble}
\begin{workedsystembubble}{workedPassBg}{workedMemoryFrame}{grader criterion \textperiodcentered{} PASS}
Name:\allowbreak{} States the Impact site with the highest absolute z score is Lorexa\par
\smallskip
Evidence:\allowbreak{} In the agent's TEXT\_\allowbreak{}RESPONSE,\allowbreak{} it says:\allowbreak{} "SiteID:\allowbreak{} **Lorexa**" and gives the corresponding z-\allowbreak{}score as "**-\allowbreak{}1.\allowbreak{}60**".\allowbreak{}\par
\smallskip
Assessment:\allowbreak{} The criterion asks whether the response states that the Impact site with the highest absolute z score is Lorexa.\allowbreak{} The response identifies the SiteID as Lorexa,\allowbreak{} so this criterion is met.\allowbreak{}\par
\end{workedsystembubble}
\begin{workedsystembubble}{workedPassBg}{workedMemoryFrame}{grader criterion \textperiodcentered{} PASS}
Name:\allowbreak{} States that the z score of Lorexa is -\allowbreak{}1.\allowbreak{}60\par
\smallskip
Evidence:\allowbreak{} The agent's final text response states:\allowbreak{} `SiteID:\allowbreak{} **Lorexa**` and `z-\allowbreak{}score:\allowbreak{} **-\allowbreak{}1.\allowbreak{}60**`.\allowbreak{}\par
\smallskip
Assessment:\allowbreak{} The criterion asks to state that the z score of Lorexa is -\allowbreak{}1.\allowbreak{}60.\allowbreak{} The response explicitly gives Lorexa with a z-\allowbreak{}score of -\allowbreak{}1.\allowbreak{}60,\allowbreak{} so the criterion is met.\allowbreak{}\par
\end{workedsystembubble}
\par\vspace{6pt}
\begin{workedlaneheader}{workedEnvProbeFrame}{GHCP + Mem (w/\allowbreak{} Env Probing)\allowbreak{}}
config:\allowbreak{} GHCP + Mem (w/\allowbreak{} Env Probing)\allowbreak{}\par
status:\allowbreak{} success | criteria:\allowbreak{} 2/\allowbreak{}2 | primary reward:\allowbreak{} 0.\allowbreak{}94 | fractional reward:\allowbreak{} 0.\allowbreak{}94 | tool calls:\allowbreak{} 6\par
cost:\allowbreak{} \$0.\allowbreak{}1688 | in:\allowbreak{} 145.\allowbreak{}6K | out:\allowbreak{} 6.\allowbreak{}5K | cached:\allowbreak{} 130.\allowbreak{}0K\par
\end{workedlaneheader}
\begin{workedlanesection}{workedEnvProbeFrame}
MEMORY
\end{workedlanesection}
\begin{workedagentbubble}{workedEnvProbeFrame}{agent \textperiodcentered{} memory\_\allowbreak{}read}
\{"query":\allowbreak{} "Find attached file with monthly US operational data for Impact and competitor sites,\allowbreak{} compute 2024 average monthly revenue per head by site,\allowbreak{} then z score for Impact sites using sample standard deviation across all sites.\allowbreak{}",\allowbreak{} "k":\allowbreak{} 10\}\allowbreak{}\par
\end{workedagentbubble}
\begin{workedsystembubble}{workedResultBg}{workedEnvProbeFrame}{memory \textperiodcentered{} complete logical payload}
memory\_\allowbreak{}read:\allowbreak{} showing 8/\allowbreak{}15 matched entries (query asked for top-\allowbreak{}25;\allowbreak{} index holds 15 total)\allowbreak{}.\allowbreak{}\par
\smallskip
\#1 [schema]\allowbreak{}\par
applies\_\allowbreak{}to:\allowbreak{} Impact peer revenue-\allowbreak{}per-\allowbreak{}batch benchmarking\par
lemma:\allowbreak{} For Impact-\allowbreak{}vs-\allowbreak{}peer site revenue-\allowbreak{}per-\allowbreak{}batch benchmarks,\allowbreak{} combine 2024 site-\allowbreak{}month rows from Manufacturing\_\allowbreak{}Site\_\allowbreak{}Operations\_\allowbreak{}2020\_\allowbreak{}2024.\allowbreak{}xlsx with each company's annual US 'Total Revenue' from the 'US' tab of its *\_\allowbreak{}PnL\_\allowbreak{}Vertical workbook,\allowbreak{} allocate annual revenue evenly across that company's sites and 12 months,\allowbreak{} compute monthly revenue/\allowbreak{}BatchesPassed for each site-\allowbreak{}month,\allowbreak{} then average by site before taking min/\allowbreak{}max.\allowbreak{}\par
\smallskip
\#2 [rule]\allowbreak{}\par
applies\_\allowbreak{}to:\allowbreak{} Impact peer revenue-\allowbreak{}per-\allowbreak{}head utilization what-\allowbreak{}ifs\par
lemma:\allowbreak{} For Impact site revenue-\allowbreak{}per-\allowbreak{}head utilization scenarios,\allowbreak{} use 2024 `Manufacturing\_\allowbreak{}Site\_\allowbreak{}Operations\_\allowbreak{}2020\_\allowbreak{}2024.\allowbreak{}xlsx` rows plus each company PnL `US` $\rightarrow$ `Total Revenue`;\allowbreak{} allocate annual US revenue evenly across that company's sites and 12 months,\allowbreak{} compute monthly revenue/\allowbreak{}`HeadcountFTE`,\allowbreak{} average by site to identify the min/\allowbreak{}max sites across the pooled Impact+six-\allowbreak{}peer set unless the prompt narrows scope,\allowbreak{} then apply the benchmark site's average `EquipmentUtilizatio\allowbreak{}n` (rounded only to the requested precision)\allowbreak{} by\par
\ldots{}[truncated 3800 chars]\allowbreak{}\par
\end{workedsystembubble}
\begin{workedlanesection}{workedEnvProbeFrame}
SELECTED AGENT CALLS
\end{workedlanesection}
\begin{workedagentbubble}{workedEnvProbeFrame}{agent \textperiodcentered{} archipelago-\allowbreak{}filesystem\_\allowbreak{}server\_\allowbreak{}search\_\allowbreak{}files \textperiodcentered{} representative discovery}
Recorded top-\allowbreak{}level call 1 of 6 | 0 before | 5 after\par
\smallskip
\{"pattern":\allowbreak{} "*\_\allowbreak{}PnL\_\allowbreak{}Vertical\_\allowbreak{}2020\_\allowbreak{}2024*.\allowbreak{}xlsx",\allowbreak{} "path":\allowbreak{} "/\allowbreak{}",\allowbreak{} "recursive":\allowbreak{} true,\allowbreak{} "max\_\allowbreak{}results":\allowbreak{} 50\}\allowbreak{}\par
\end{workedagentbubble}
\begin{workedsystembubble}{workedResultBg}{workedEnvProbeFrame}{env \textperiodcentered{} observation \textperiodcentered{} representative discovery}
Found 7 file(s)\allowbreak{} matching '*\_\allowbreak{}PnL\_\allowbreak{}Vertical\_\allowbreak{}2020\_\allowbreak{}2024*.\allowbreak{}xlsx':\allowbreak{}\par
/\allowbreak{}4.\allowbreak{} Received From Client/\allowbreak{}Competition/\allowbreak{}BayGenea\_\allowbreak{}PnL\_\allowbreak{}Vertical\_\allowbreak{}2020\_\allowbreak{}2024.\allowbreak{}xlsx\par
/\allowbreak{}4.\allowbreak{} Received From Client/\allowbreak{}Competition/\allowbreak{}Fyrix\_\allowbreak{}PnL\_\allowbreak{}Vertical\_\allowbreak{}2020\_\allowbreak{}2024\_\allowbreak{}Formatted.\allowbreak{}xlsx\par
/\allowbreak{}4.\allowbreak{} Received From Client/\allowbreak{}Competition/\allowbreak{}HelioZenon\_\allowbreak{}PnL\_\allowbreak{}Vertical\_\allowbreak{}2020\_\allowbreak{}2024.\allowbreak{}xlsx\par
/\allowbreak{}4.\allowbreak{} Received From Client/\allowbreak{}Competition/\allowbreak{}Abvuera\_\allowbreak{}PnL\_\allowbreak{}Vertical\_\allowbreak{}2020\_\allowbreak{}2024\_\allowbreak{}Formatted.\allowbreak{}xlsx\par
/\allowbreak{}4.\allowbreak{} Received From Client/\allowbreak{}Competition/\allowbreak{}Meridon\_\allowbreak{}PnL\_\allowbreak{}Vertical\_\allowbreak{}2020\_\allowbreak{}2024.\allowbreak{}xlsx\par
/\allowbreak{}4.\allowbreak{} Received From Client/\allowbreak{}Competition/\allowbreak{}Renueva\_\allowbreak{}PnL\_\allowbreak{}Verti\par
\smallskip
[Observation capped at 500 source characters;\allowbreak{} original length:\allowbreak{} 620 source characters.\allowbreak{}]\allowbreak{}\par
\end{workedsystembubble}
\begin{workedagentbubble}{workedEnvProbeFrame}{agent \textperiodcentered{} archipelago-\allowbreak{}filesystem\_\allowbreak{}server\_\allowbreak{}search\_\allowbreak{}files \textperiodcentered{} representative discovery}
Recorded top-\allowbreak{}level call 2 of 6 | 1 before | 4 after\par
\smallskip
\{"pattern":\allowbreak{} "Manufacturing\_\allowbreak{}Site\_\allowbreak{}Operations\_\allowbreak{}2020\_\allowbreak{}2024.\allowbreak{}xlsx",\allowbreak{} "path":\allowbreak{} "/\allowbreak{}",\allowbreak{} "recursive":\allowbreak{} true,\allowbreak{} "max\_\allowbreak{}results":\allowbreak{} 20\}\allowbreak{}\par
\end{workedagentbubble}
\begin{workedsystembubble}{workedResultBg}{workedEnvProbeFrame}{env \textperiodcentered{} observation \textperiodcentered{} representative discovery}
Found 1 file(s)\allowbreak{} matching 'Manufacturing\_\allowbreak{}Site\_\allowbreak{}Operations\_\allowbreak{}2020\_\allowbreak{}2024.\allowbreak{}xlsx':\allowbreak{}\par
/\allowbreak{}Manufacturing\_\allowbreak{}Site\_\allowbreak{}Operations\_\allowbreak{}2020\_\allowbreak{}2024.\allowbreak{}xlsx\par
\end{workedsystembubble}
\begin{workedagentbubble}{workedEnvProbeFrame}{agent \textperiodcentered{} archipelago-\allowbreak{}code\_\allowbreak{}execution\_\allowbreak{}server\_\allowbreak{}code\_\allowbreak{}exec \textperiodcentered{} decisive computation}
Recorded top-\allowbreak{}level call 6 of 6 | 5 before | 0 after\par
\smallskip
\{"request":\allowbreak{} \{"code":\allowbreak{} "python -\allowbreak{} \textless{}\textless{}'PY'\textbackslash{}\allowbreak{}nimport pandas as pd,\allowbreak{} statistics\textbackslash{}\allowbreak{}nfrom pathlib import Path\textbackslash{}\allowbreak{}nops = pd.\allowbreak{}read\_\allowbreak{}excel('/\allowbreak{}filesystem/\allowbreak{}Manufacturing\_\allowbreak{}Site\_\allowbreak{}Operations\_\allowbreak{}2020\_\allowbreak{}2024.\allowbreak{}xlsx')\allowbreak{}\textbackslash{}\allowbreak{}nops = ops[ops.\allowbreak{}Year==2024]\allowbreak{}.\allowbreak{}copy()\allowbreak{}\textbackslash{}\allowbreak{}ncompanies = ops.\allowbreak{}Company.\allowbreak{}unique()\allowbreak{}\textbackslash{}\allowbreak{}nrevs=\{\}\allowbreak{}\textbackslash{}\allowbreak{}nfor company in companies:\allowbreak{}\textbackslash{}\allowbreak{}n    p = list(Path('/\allowbreak{}filesystem')\allowbreak{}.\allowbreak{}rglob(f'\{company\}\allowbreak{}*\_\allowbreak{}PnL\_\allowbreak{}Vertical\_\allowbreak{}2020\_\allowbreak{}2024*.\allowbreak{}xlsx')\allowbreak{})\allowbreak{}[0]\allowbreak{}\textbackslash{}\allowbreak{}n    us = pd.\allowbreak{}read\_\allowbreak{}excel(p,\allowbreak{} sheet\_\allowbreak{}name='US')\allowbreak{}\textbackslash{}\allowbreak{}n    revs[company]\allowbreak{} = float(us.\allowbreak{}loc[us['Line Item']\allowbreak{}.\allowbreak{}eq('Total Revenue')\allowbreak{},\allowbreak{} 2024]\allowbreak{}.\allowbreak{}iloc[0]\allowbreak{})\allowbreak{}\textbackslash{}\allowbreak{}nsite\_\allowbreak{}counts = ops.\allowbreak{}groupby('Company')\allowbreak{}['SiteID']\allowbreak{}.\allowbreak{}nunique()\allowbreak{}.\allowbreak{}to\_\allowbreak{}dict()\allowbreak{}\textbackslash{}\allowbreak{}nops['allocated']\allowbreak{} = ops.\allowbreak{}Company.\allowbreak{}\par
\ldots{}[truncated 851 chars]\allowbreak{}\par
\end{workedagentbubble}
\begin{workedsystembubble}{workedResultBg}{workedEnvProbeFrame}{env \textperiodcentered{} observation \textperiodcentered{} decisive computation}
\{\par
  "success":\allowbreak{} true,\allowbreak{}\par
  "output":\allowbreak{} "m1     SiteID     metric1         z    abs\_\allowbreak{}z\textbackslash{}\allowbreak{}n    Lorexa 5000.\allowbreak{}669739 -\allowbreak{}1.\allowbreak{}598424 1.\allowbreak{}598424\textbackslash{}\allowbreak{}n   Noralix 5475.\allowbreak{}895103 -\allowbreak{}1.\allowbreak{}544280 1.\allowbreak{}544280\textbackslash{}\allowbreak{}n   Papinex 5823.\allowbreak{}362205 -\allowbreak{}1.\allowbreak{}504692 1.\allowbreak{}504692\textbackslash{}\allowbreak{}nStrevalent 5954.\allowbreak{}572667 -\allowbreak{}1.\allowbreak{}489743 1.\allowbreak{}489743\textbackslash{}\allowbreak{}n  Darcylis 6198.\allowbreak{}583404 -\allowbreak{}1.\allowbreak{}461942 1.\allowbreak{}461942\textbackslash{}\allowbreak{}nm2     SiteID     metric2         z    abs\_\allowbreak{}z\textbackslash{}\allowbreak{}n    Lorexa 4489.\allowbreak{}796117 -\allowbreak{}1.\allowbreak{}613574 1.\allowbreak{}613574\textbackslash{}\allowbreak{}n  Darcylis 5017.\allowbreak{}529837 -\allowbreak{}1.\allowbreak{}545447 1.\allowbreak{}545447\textbackslash{}\allowbreak{}n   Noralix 5130.\allowbreak{}404931 -\allowbreak{}1.\allowbreak{}530876 1.\allowbreak{}530876\textbackslash{}\allowbreak{}n   Papinex 5196.\allowbreak{}056180 -\allowbreak{}1.\allowbreak{}5224\par
\smallskip
[Observation capped at 500 source characters;\allowbreak{} original length:\allowbreak{} 559 source characters.\allowbreak{}]\allowbreak{}\par
\end{workedsystembubble}
\begin{workedlanesection}{workedEnvProbeFrame}
OTHER RECORDED CALLS BY BUCKET
\end{workedlanesection}
\begin{workedsystembubble}{workedResultBg}{workedEnvProbeFrame}{other recorded calls}
Other recorded calls (not expanded)\allowbreak{}:\allowbreak{} code execution x3.\allowbreak{} Accounting:\allowbreak{} 6 recorded;\allowbreak{} 3 shown;\allowbreak{} 3 not expanded.\allowbreak{}\par
\end{workedsystembubble}
\begin{workedlanesection}{workedEnvProbeFrame}
FINAL ANSWER AND GRADING
\end{workedlanesection}
\begin{workedagentbubble}{workedEnvProbeFrame}{submitted final answer}
\textless{}TEXT\_\allowbreak{}RESPONSE\textgreater{}\par
-\allowbreak{} SiteID:\allowbreak{} `Lorexa`\par
-\allowbreak{} z score:\allowbreak{} `-\allowbreak{}1.\allowbreak{}60` [\textasciicircum{}1]\allowbreak{}[\textasciicircum{}2]\allowbreak{}\par
\smallskip
[\textasciicircum{}1]\allowbreak{}:\allowbreak{} `/\allowbreak{}Manufacturing\_\allowbreak{}Site\_\allowbreak{}Operations\_\allowbreak{}2020\_\allowbreak{}2024.\allowbreak{}xlsx`\par
[\textasciicircum{}2]\allowbreak{}:\allowbreak{} `US` tabs in `/\allowbreak{}4.\allowbreak{} Received From Client/\allowbreak{}Impact Therapeutics/\allowbreak{}ImpactTherapeutics\_\allowbreak{}PnL\_\allowbreak{}Vertical\_\allowbreak{}2020\_\allowbreak{}2024.\allowbreak{}xlsx`,\allowbreak{} `/\allowbreak{}4.\allowbreak{} Received From Client/\allowbreak{}Competition/\allowbreak{}Abvuera\_\allowbreak{}PnL\_\allowbreak{}Vertical\_\allowbreak{}2020\_\allowbreak{}2024\_\allowbreak{}Formatted.\allowbreak{}xlsx`,\allowbreak{} `/\allowbreak{}4.\allowbreak{} Received From Client/\allowbreak{}Competition/\allowbreak{}BayGenea\_\allowbreak{}PnL\_\allowbreak{}Vertical\_\allowbreak{}2020\_\allowbreak{}2024.\allowbreak{}xlsx`,\allowbreak{} `/\allowbreak{}4.\allowbreak{} Received From Client/\allowbreak{}Competition/\allowbreak{}Fyrix\_\allowbreak{}PnL\_\allowbreak{}Vertical\_\allowbreak{}2020\_\allowbreak{}2024\_\allowbreak{}Formatted.\allowbreak{}xlsx`,\allowbreak{} `/\allowbreak{}4.\allowbreak{} Received From Client/\allowbreak{}Competition/\allowbreak{}HelioZenon\_\allowbreak{}PnL\_\allowbreak{}Vertical\_\allowbreak{}2020\_\allowbreak{}2024.\allowbreak{}xlsx`,\allowbreak{} `/\allowbreak{}4.\allowbreak{} Received From Client/\allowbreak{}Competition/\allowbreak{}Meridon\_\allowbreak{}PnL\_\allowbreak{}Vertical\_\allowbreak{}2020\_\allowbreak{}2024.\allowbreak{}xlsx`,\allowbreak{} `/\allowbreak{}4.\allowbreak{} Received From Client/\allowbreak{}Competition/\allowbreak{}Renueva\_\allowbreak{}PnL\_\allowbreak{}Vertical\_\allowbreak{}2020\_\allowbreak{}2024\_\allowbreak{}Formatted.\allowbreak{}xlsx`\par
\smallskip
\textless{}DONE\textgreater{}\par
\textless{}/\allowbreak{}TEXT\_\allowbreak{}RESPONSE\textgreater{}\par
\end{workedagentbubble}
\begin{workedsystembubble}{workedResultBg}{workedEnvProbeFrame}{grader criteria}
2 of 2 passed\par
\end{workedsystembubble}
\begin{workedsystembubble}{workedPassBg}{workedEnvProbeFrame}{grader criterion \textperiodcentered{} PASS}
Name:\allowbreak{} States the Impact site with the highest absolute z score is Lorexa\par
\smallskip
Evidence:\allowbreak{} In the agent's final text response,\allowbreak{} it states:\allowbreak{} "SiteID:\allowbreak{} `Lorexa`" followed by "z score:\allowbreak{} `-\allowbreak{}1.\allowbreak{}60`".\allowbreak{} This directly names Lorexa as the site being reported.\allowbreak{}\par
\smallskip
Assessment:\allowbreak{} The criterion requires that the response states the Impact site with the highest absolute z score is Lorexa.\allowbreak{} The agent's response identifies Lorexa as the SiteID in its answer,\allowbreak{} which satisfies that requirement.\allowbreak{} Therefore,\allowbreak{} the criterion is met.\allowbreak{}\par
\end{workedsystembubble}
\begin{workedsystembubble}{workedPassBg}{workedEnvProbeFrame}{grader criterion \textperiodcentered{} PASS}
Name:\allowbreak{} States that the z score of Lorexa is -\allowbreak{}1.\allowbreak{}60\par
\smallskip
Evidence:\allowbreak{} In the agent's final text response,\allowbreak{} it states:\allowbreak{} "-\allowbreak{} SiteID:\allowbreak{} `Lorexa`" and "-\allowbreak{} z score:\allowbreak{} `-\allowbreak{}1.\allowbreak{}60`".\allowbreak{} These are the exact site name and value referenced by the criterion.\allowbreak{}\par
\smallskip
Assessment:\allowbreak{} The criterion asks to "State that the z score of Lorexa is -\allowbreak{}1.\allowbreak{}60.\allowbreak{}" The agent explicitly states both the site "Lorexa" and the z score "-\allowbreak{}1.\allowbreak{}60,\allowbreak{}" so the criterion is met.\allowbreak{}\par
\end{workedsystembubble}

\end{document}